\documentclass{article}

\usepackage[preprint]{neurips_2023}

\usepackage{microtype}
\usepackage{graphicx}
\usepackage{subfig}
\usepackage{booktabs} 
\usepackage{enumitem}

\usepackage{hyperref}

\usepackage{amsmath}
\usepackage{amssymb}
\usepackage{mathtools}
\usepackage{amsthm}

\usepackage[capitalize,noabbrev]{cleveref}

\theoremstyle{plain}

\theoremstyle{definition}

\theoremstyle{remark}

\newcommand{\ignore}[1]{}

\newenvironment{squishitem}
    {\begin{itemize}[itemsep=0pt, parsep=3pt, topsep=3pt, partopsep=5pt, leftmargin=1em, labelwidth=1em, labelsep=0.5em]}
    {\end{itemize}}

\usepackage[textsize=tiny]{todonotes}

\usepackage[utf8]{inputenc} 
\usepackage[T1]{fontenc}    
\usepackage{hyperref}       
\usepackage{url}            
\usepackage{booktabs}       
\usepackage{amsfonts}       
\usepackage{nicefrac}       
\usepackage{microtype}      
\usepackage{xcolor}         

\title{Substance or Style: What Does Your Image Embedding Know?}

\author{%
  Cyrus Rashtchian\thanks{Equal contribution.}\\
  Google Research\\
  \texttt{cyroid@google.com} \\
  \And
  Charles Herrmann$^*$\\
  Google Research\\
  \texttt{irwinherrmann@google.com} \\
  \And
  Chun-Sung Ferng$^*$\\
  Google Research\\
  \texttt{csferng@google.com} \\
  \And
  Ayan Chakrabarti\\
  Google Research\\
  \texttt{ayanchakrab@google.com} \\
  \And
  Dilip Krishnan\\
  Google Research\\
  \texttt{dilipkay@google.com} \\
  \And
  Deqing Sun\\
  Google Research\\
  \texttt{deqingsun@google.com} \\
  \And
  Da-Cheng Juan\\
  Google Research\\
  \texttt{dacheng@google.com} \\
  \And
  Andrew Tomkins\\
  Google Research\\
  \texttt{tomkins@google.com} \\
}

\begin{document}

\maketitle

\begin{abstract}
Probes are small networks that predict properties of underlying data from embeddings, and they provide a targeted, effective way to illuminate the information contained in embeddings. While analysis through the use of probes has become standard in NLP, there has been much less exploration in vision. Image foundation models have primarily been evaluated for semantic content. Better understanding the non-semantic information in popular embeddings (e.g., MAE, SimCLR, or CLIP) will shed new light both on the training algorithms and on the uses for these foundation models. We design a systematic transformation prediction task and measure the visual content of embeddings along many axes, including image style, quality, and a range of natural and artificial transformations. Surprisingly, six embeddings (including SimCLR) encode enough non-semantic information to identify dozens of transformations. We also consider a generalization task, where we group similar transformations and hold out several for testing. We find that image-text models (CLIP and ALIGN) are better at recognizing new examples of style transfer than masking-based models (CAN and MAE). Overall, our results suggest that the choice of pre-training algorithm impacts the types of information in the embedding, and certain models are better than others for non-semantic downstream tasks.
\end{abstract}

\section{Introduction}
\label{intro}

Large pre-trained networks, sometimes known as \emph{foundation models}, provide a `general-purpose’ embedding for multiple data modalities~\citep{bommasani2021opportunities, zhou2023comprehensive}. The models perform very well on several downstream tasks and exhibit robustness to dataset shift. In large ML systems, raw data is often pre-processed using embeddings, and training a small network on top of a frozen embedding is a scalable and desirable solution.  A central research direction is to develop foundation models that 
are easily adapted for current and future applications. Hence, it is important to be able to evaluate what information these embeddings capture and what aspects they ignore.

\begin{figure*}[ht]
\centering
\includegraphics[width=0.8\textwidth]{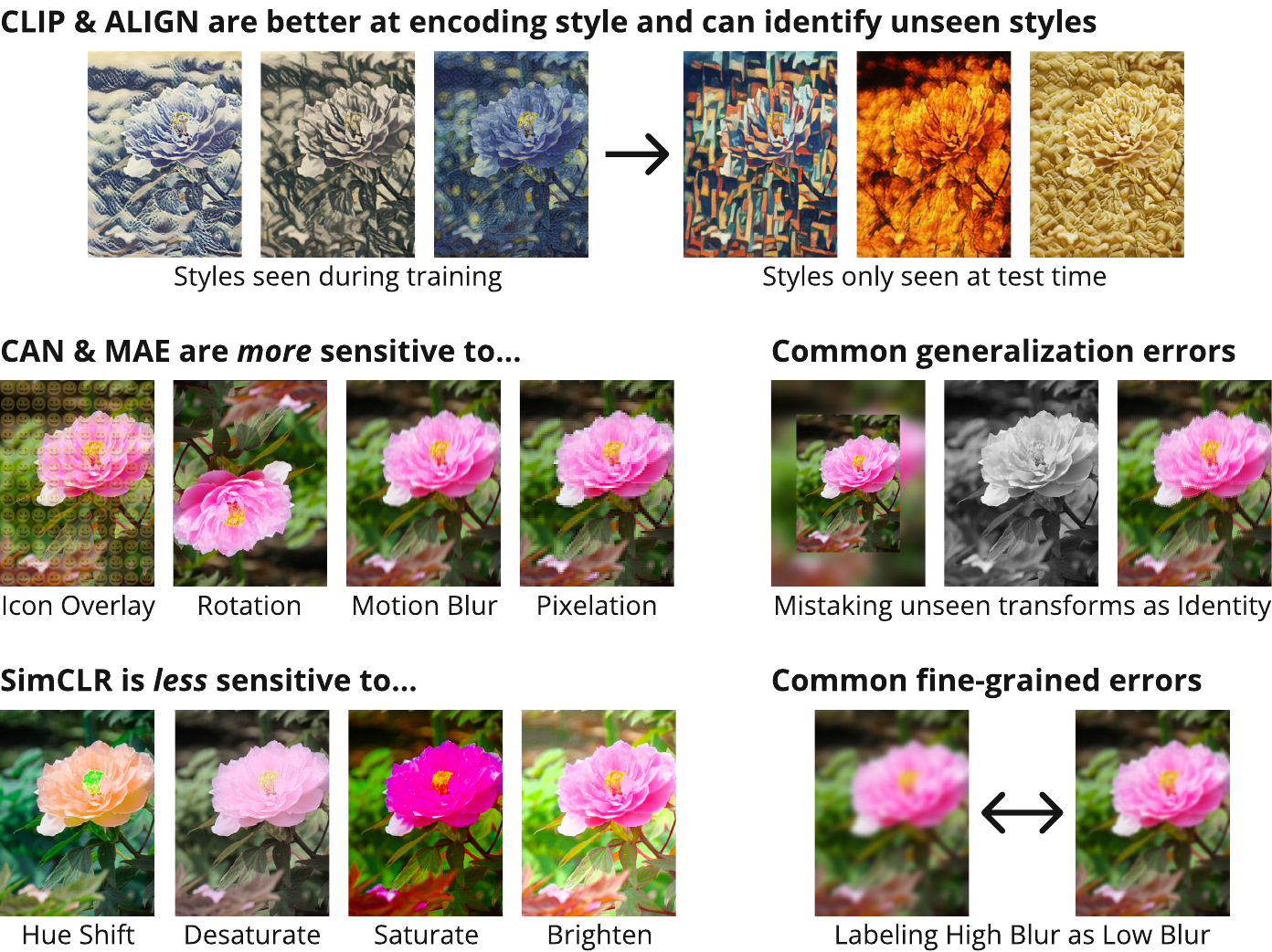}
\caption{Main takeaways from analyzing the performance of frozen image embeddings on our transformation prediction tasks. We draw conclusions about (in)sensitivity by looking the accuracy when detecting whether a particular transformation has been applied to an image. We evaluate both a fine-grained version (31 classes, same train and test transforms) and a generalization version (10 classes, with 28 training and 15 additional test transforms grouped into categories). Both versions contain the `Identity' class as the original image.}
\label{fig:summary}
\end{figure*}

Probes provide a way to determine what information can be extracted about data after computing an embedding. The idea is to train a small network (a.k.a., probe) to predict certain properties about the underlying data using only an embedding of the data~\citep{alain2016understanding}. 
In addition to helping explain embedding models, probes also  inform whether embeddings can be used for downstream tasks that rely on certain information about the data. In NLP, researchers have probed text embeddings for 
information about syntax trees, sentence length, word order, and so on~\citep{belinkov2021probing, conneau2018you,hewitt2019structural, li2022emergent}. For image-text models, it is also possible to probe the visual information through text prompts or questions that describe visual attributes, such as color, shape, and material~\citep{liu2022things, paik-etal-2021-world, zhang2022visual}. 

Unfortunately, probes using text prompts are limited to concepts describable in short text snippets. This overlooks visual attributes that are important for certain tasks. For example, a core part of trust and safety involves filtering content for policy violations, which is challenging when adversaries manipulate images to evade filters~\citep{cao2022stylefool,goodman2019cloud,hao2021s,li2019adversarial,stimberg2023benchmarking, yuan_2019_stealthy}. These manipulations include noise, recoloring, overlays, and style transfer. One solution is to develop filtering models that are highly invariant. However, it is also critical to produce manipulation \emph{detectors}, to identify attempts to bypass filters and intervene at the user level. Unlike semantic classification, the new task of manipulation detection must instead use embeddings that are highly sensitive to the transformations that we wish to detect.

Can a single embedding be robust against dataset shift, and yet, sensitive to many transformations? And even if it is possible, are today's popular pre-trained embeddings successful at doing so? These are the questions we study in this paper. Through image probing experiments, we evaluate the sensitivity of several popular embeddings to dozens of transformations (see \cref{fig:summary} for examples) and offer insights into the information encoded by image-only and image-text foundation models.

\subsection{Visual Probing Using Image Transformations}

Our new approach to visual probing centers around predicting how an image has been transformed. 
The core idea of the experiment is to modify an image and then see if this change is detectable after computing the image’s embedding. For example, consider two images: one that is a sample from ImageNet, and another where the same image has been slightly blurred. Then, compute embeddings for both images and throw out the original pixels. Assume that in both cases a linear probe will predict the correct ImageNet class for the image. The next question is: does the embedding contain enough information to determine which image was blurred and which was unaltered? 

If the embedding contains sufficient information to detect blurring, then it should be possible to train a network to perform well on a `blurry or not’ classification task. Specifically, we can apply Gaussian blur to all images in ImageNet, and we can train a probing network to predict whether the transformation has been applied given access only to the image embeddings. Foundation models that capture more of the transformation information will perform better on this task, whereas models that perform poorly must be insensitive to the transformation. Note that freezing the embedding model is crucial for this analysis. If we fine-tuned on the transformation prediction task, then we would not know whether the original model extracted the transformation information or not.

One of our goals is to interpret the information that is kept or lost when using popular vision embeddings. Another goal is to evaluate the effectiveness of embeddings for the task of predicting transformations. Our motivation comes from using embeddings for tasks such as detecting manipulation, predicting quality, or data cleaning. We carefully design the set of transformations, ensuring enough variety to elicit whether embeddings capture different types of visual content. For example, we include image filtering (e.g, hue shift, saturate), occlusion (e.g., icon or text overlay), corruption (e.g., noise, pixelate), natural domain shift (e.g., motion blur, brighten, crop), and style transfer.

Using our set of transformations, we propose two transformation prediction tasks. The first is a \emph{fine-grained} variation. Here, the probe sees all 30 transformations during training and the goal is to classify them. We design the fine-grained transformations so that we expect trained humans to achieve 100\% accuracy. Any errors from the probe indicate a lack of information in the embedding. 

Our second task focuses on \emph{generalization}. We group the transformations into 9 categories, and we only train the probe using some in each category.  During test time, the probe should recognize the category of the transformation. This task measures two things: (1) whether an embedding organizes transformation information in a generalizable way, and (2) whether we can use the embedding for more realistic prediction with held-out transformations. 
Specifically, for (2), we may want to detect domain shifts, such as image manipulations or natural perturbations. However, we cannot train with all relevant transformations. Instead, we desire a model that can correctly classify unseen, but similar, transformations. For example, we have a category based on style transfer, but the probe only sees some styles during training. Nonetheless, transferring other styles still produces image that have been modified in a visually analogous way. Our generalization task evaluates whether the features in the embedding are sufficient to also detect new examples in the same transformation category.

We evaluate a representative sample of vision foundation models based on their pre-training algorithms: (1) a masked autoencoder (MAE)~\citep{he2022masked}, a canonical masking-based method that fills image portions during pre-training, (2) SimCLR~\citep{chen2020simple}, an image-only contrastive loss that encourages invariance to a few transformations, (3) CAN~\citep{mishra2022simple}, which combines masking, contrastive, and noise prediction, (4) CLIP~\citep{radford2021learning}, a standard image-text self-supervised method, (5) ALIGN~\citep{jia2021scaling}, another contrastive image-text model, and (6) a supervised method that trains with the semantic class labels.


\subsection{Our Contributions}

Our image embedding probing tasks lead to many new insights.  Image-text models, CLIP and ALIGN, excel in recognizing the concept of style transfer. Probes trained with them can generalize to detect new styles after being only trained on a few. These models are also consistently sensitive to all of our fine-grained transformations. This is surprising since many transformations do not have simple text descriptions, yet the model's features retain the relevant information.

Image-only contrastive learning algorithms have quite different dispositions, even when trained on the same, large dataset. CAN and MAE are more sensitive than SimCLR. Specifically, SimCLR is fairly invariant to hue, saturation, and brightness, but it is sensitive to other transformations (including blurring, which is part of the contrastive training). Our supervised baseline encodes enough information for the fine-grained task, but does not generalize well to unseen transformations.

Beyond probing, another natural question is whether post-processing the embedding to improve transformation prediction will effect the semantic accuracy (e.g., ImageNet top-1 accuracy). We actually find that it is possible to achieve good performance on both metrics when training a 2-layer MLP with two heads and optimizing a multi-task loss. This implies that the transformation information does not interfere with the object-level features. Both can coexist.

With transformed images there is a related question around robust accuracy. Common wisdom suggests that generalizing to OOD data can be facilitated by encouraging invariance to corruptions or style~\citep{ganin2016domain, mao2022enhance}. 
Instead, we find that embeddings with a lot of transformation content do not perform worse in terms of data shift robustness. 
In general, the sensitivity of foundation models implies that their robustness to domain shift is not due to invariance. 

In summary, our main findings are (see also \cref{fig:summary}):
\begin{squishitem}
    \item Foundation models are sensitive to and encode information about dozens of transformations. Hence, we can use embeddings to detect a domain shift due to transformations (e.g., image manipulation).
    \item Vision models with masking (CAN, MAE) are more sensitive than those using only a contrastive loss (SimCLR)
    to changes in hue, saturation, and brightness.
    \item 
    Image-text models (CLIP and ALIGN) encode transformation information in a more generalizable way than image-only embeddings, especially for style transfer. 
    Probes using image-only embeddings perform worse than chance level for several unseen styles.
    \item Many errors in the generalization task come from mistaking images as normal (i.e., `Identity' transform) when they have been modified in unseen ways (e.g., background blur, grayscale, line shift). Since these errors are more common for some held-out transformations than others, there is room to improve the information in the embeddings.
    \item Transformation prediction fine-tunes embeddings to surface non-semantic content, and sharing a hidden layer for semantic and transformation prediction does not harm performance on either task. 
\end{squishitem}

\section{Probing Embeddings by Predicting Transformations}
\label{probing}

Evaluating only the typical semantic accuracy on class labels leaves open questions regarding what information from the raw data is retained or lost in the embedding. Instead, we probe the embeddings, measuring the ability of a network to determine how an image has been transformed. To do so, we define a transformation prediction task along with new metrics. This task can be formulated for any dataset/task as long as there is a way to synthetically apply transformations.

\begin{figure*}[ht]
\centering
\subfloat[Identity\label{fig:1a}]{\includegraphics[width=0.162\textwidth]{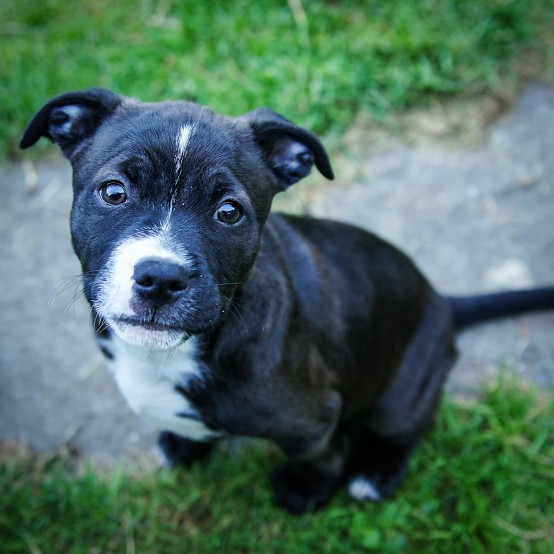}}\hfill
\subfloat[Line Shift\label{fig:1b}]{\includegraphics[width=0.162\textwidth]{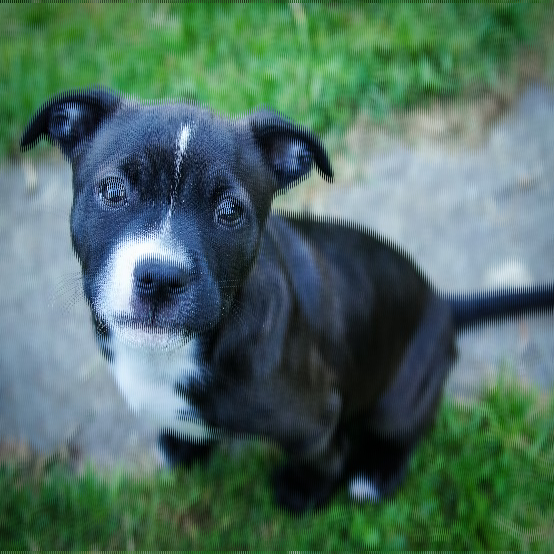}}\hfill
\subfloat[Motion Blur\label{fig:1c}]{\includegraphics[width=0.162\textwidth]{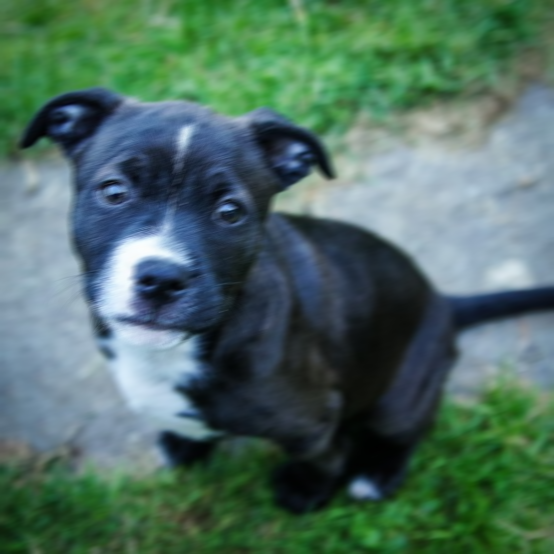}}\hfill
\subfloat[Low Noise\label{fig:1d}]
{\includegraphics[width=0.162\textwidth]{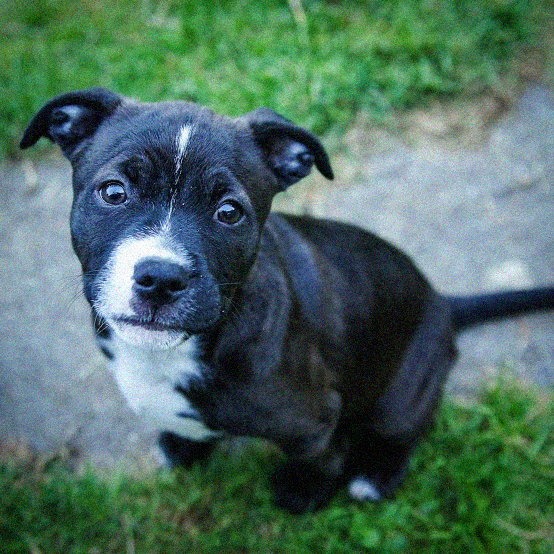}}\hfill
\subfloat[Medium Noise\label{fig:1e}] {\includegraphics[width=0.162\textwidth]{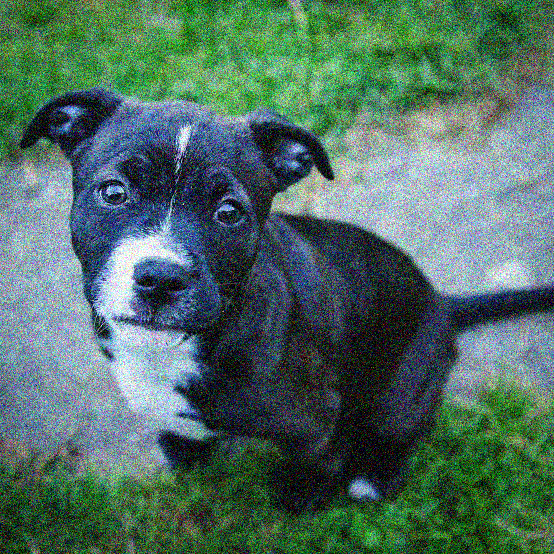}}\hfill
%
\subfloat[Darken\label{fig:1f}]{\includegraphics[width=0.162\textwidth]{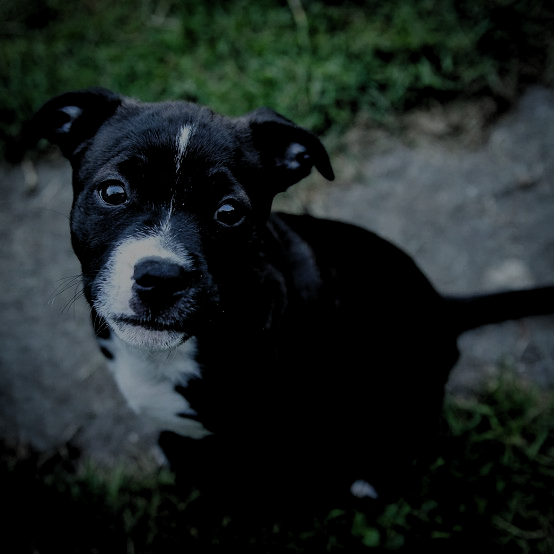}}\hfill
\subfloat[Hue Rescale\label{fig:1g}] {\includegraphics[width=0.162\textwidth]{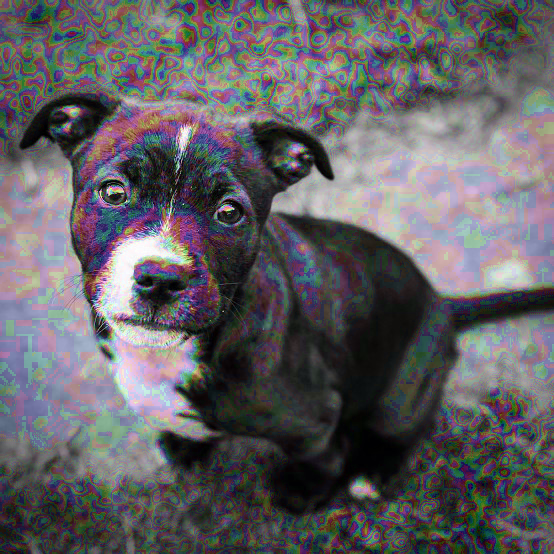}}\hfill
\subfloat[Posterize\label{fig:1h}] {\includegraphics[width=0.162\textwidth]{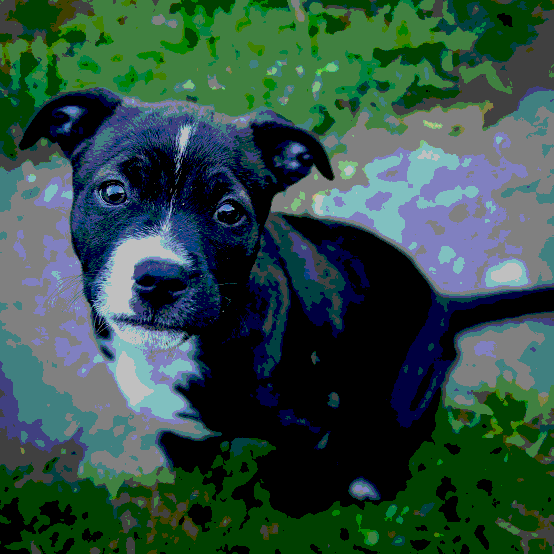}}\hfill
\subfloat[Solarize\label{fig:1i}]{\includegraphics[width=0.162\textwidth]{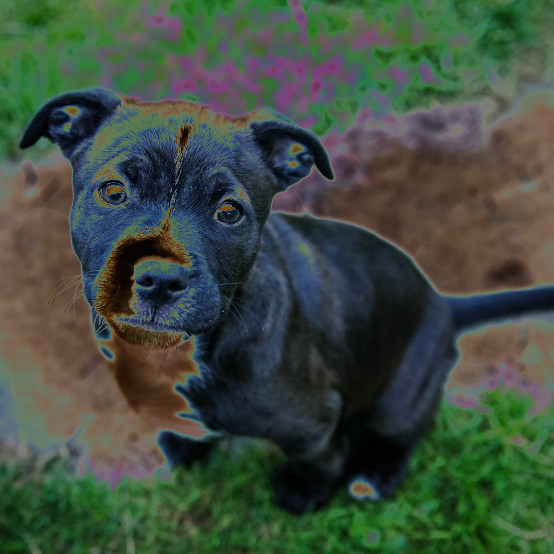}}\hfill
\subfloat[Halftoning\label{fig:1j}] {\includegraphics[width=0.162\textwidth]{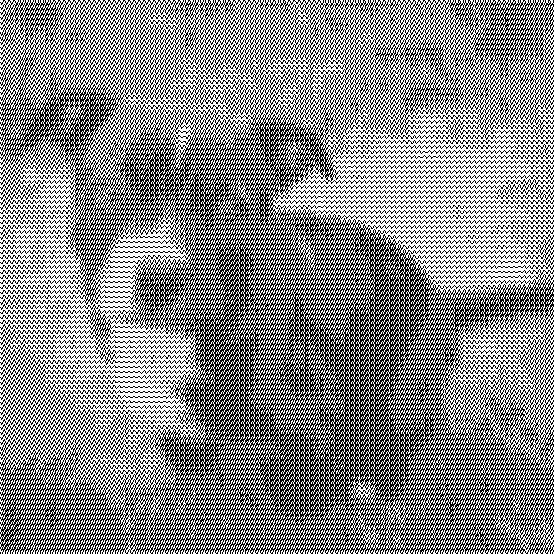}}\hfill
\subfloat[Starry Night\label{fig:1k}]{\includegraphics[width=0.162\textwidth]{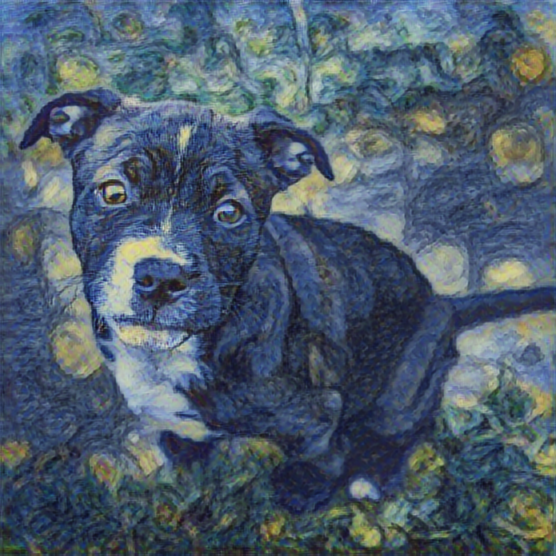}}\hfill
\subfloat[Bonfire\label{fig:1l}] {\includegraphics[width=0.162\textwidth]{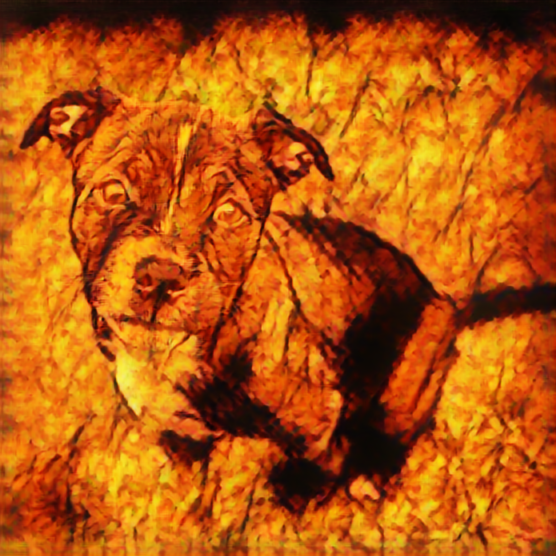}}\hfill
\caption{Example transformations that we train a network to predict from an embedding. They contain a mix of effects, altering all or part of the image. This enables us to test for different information in the embeddings. In our experiments, we evaluate dozens of transformations, and for many, we randomize the parameters for each image. We also consider a generalization setting where the network is only trained on a subset of transformations and must make predictions for images with unseen transformations from the same category (e.g., different styles).
}
\label{fig:transformations}
\end{figure*}

\label{metrics}

Assume we have $T$ image transformations (\cref{fig:transformations} shows examples). Here, for \emph{transformation}, we take a broad definition. One option is a well-defined function, such as adding Gaussian noise with certain variance independently to each pixel. Another possibility is to have some random parameters, such as uniformly choosing a value in a range and increasing the image’s saturation by this much. Finally, we can have transformation families, containing several \emph{sub-transformations}. For example, the family ``color quantizing’’ could mean choosing a sub-transformation that modifies hue, inverts colors, or solarizes the image. Sub-transformations have their own (possibly random) parameters.

We apply each of the $T$ transformations to all images in the training/test sets. This generates $T+1$ copies of the dataset, including the original images. Also, this process defines a $(T+1)$-way classification problem, labeling each image either with `Identity’ or one of the $T$ transformations.

\textbf{Metrics.} Our tasks involve both unaltered (clean) images and transformed ones, as well as a new label for the type of transformation. For a dataset such as ImageNet, which contains images $x$ and semantic class labels $y$, we will use $t$ to denote the transformation label of our augmented dataset, leading to a labeled triple $(x,y,t)$. A network can predict the semantic label $y$, the transformation label $t$, or both in a multi-task scenario. 
The \emph{transformation prediction accuracy} is the fraction of images receiving the correct transformation label (the network does not see the class label). We use \emph{clean semantic accuracy} to refer to the fraction of correctly predicted class labels on unaltered images (i.e., the transformation $t$ is the identity). The \emph{obfuscated semantic accuracy} is the fraction of correct class labels when the image has been transformed (i.e., $t$ is \emph{not} the identity).

\subsection{Probing Frozen Image Embeddings}

Consider an image $x$, let $t$ be one of the $T+1$ transformations, and use $t(x)$ to denote the transformed version of $x$. For a frozen embedding model $\phi$, we compute the embedding $\phi(t(x))$. We then train the probe, which is a small network that takes $\phi(t(x))$ as input and outputs a semantic label or a transformation label or both. In a multi-task setting with a two-headed network that outputs two labels, we independently measure the clean/obfuscated semantic and transformation accuracies.

Training a linear probe on top of the embedding $\phi(t(x))$ is the simplest setting to probe embeddings~\citep{alain2016understanding}. The probe weights are trained using the transformation labels. We are also interested in post-processing the representation to extract more information about transformations. In this case, we train an MLP with a single hidden layer instead of a linear probe. This leads to higher transformation prediction accuracy. More importantly, we can extend it to a multi-task setting. We combine the loss from both the semantic and transformation prediction tasks during training (using a two-headed network). Doing so leads to a post-processed representation (i.e., the hidden layer) with both semantic and transformation information. Moreover, we can achieve this combination with only light-weight training of the MLP, instead of fine-tuning the embedding model. 

Freezing the embedding model has another benefit: we do not over-process the embedding. If we fine-tuned the whole embedding model, then we would conflate the information in the original embedding with the new information learned from the fine-tuning. Instead, any information in the embedding must have been there to begin with because it cannot be extracted from the raw pixels. \cref{sec:drawing-conclusions} further explains why our experiments illuminate the sensitivities of embeddings.



\subsection{Two Image Probing Tasks: Fine-grained vs.~Generalization}

We consider a fine-grained task (train and test sets use the same transformations) and a generalization version (the same label may contain different sub-transformations). For both tasks, the post-processing network learns which features of the embedding correspond to the transformations.

The fine-grained task measures the granularity of information in embeddings. It has 31 labels, including `Identity' for unaltered images. In a few cases, we use the same transformation with disjoint parameter ranges as separate classes. Two categories come from each of (i) a low or medium amount of motion blur, (ii) a low or high amount of Gaussian blur, (iii) a low or medium amount of Gaussian noise, and (iv) increasing or decreasing the brightness. At test-time, the same transformation applied to an image will only differ in its randomized parameters. Given the visibility of the transformations, we expect 100\% accuracy to be feasible, with errors due to embedding insensitivity.

The generalization task measures whether embeddings encode similar transformations in a way that allows the probe to identify that they come from the same category. It also mimics real-world usage for detecting specific OOD data, e.g., perceptibly manipulated images~\citep{stimberg2023benchmarking}. The training set has 28 transformations, split across 9 categories, plus the Identity. The test set has 43 sub-transformations, split across the same categories.
There are 15 held-out transformations that the network only sees during testing. We define the coarse categories so that the visual content should be similar in some way. For example, `Quantize' contains 7 recoloring options (4 for training and 3 held-out).  The `Style Transfer' label has 13 style options (6 for training and 7 held-out). Three categories (Icon Overlay, Image Overlay, Halftoning) have no held-out sub-transformations. Full details are in \cref{dataset-details}, with more motivation for transformation choices in \cref{sec:transformation-motivation}.

\section{Experimental Results}
\label{experiments}

\textbf{Datasets.} We evaluate with transformed versions of ImageNet-1k~\citep{russakovsky_2015_imagenet}. In addition to the original image (Identity), we apply 30 transformations to each train/test image. This leads to 31 classes for the fine-grained transformation prediction task. We also construct a generalization dataset with 10 categories, where each category contains one or more transformations along with a range of parameters (e.g., noise level or type of style transfer). The test set transformations form a superset of those applied to the training images. Full details in \cref{dataset-details}. 

\textbf{Metrics}. We measure semantic and transformation prediction accuracies as defined in \cref{metrics}. In the fine-grained case, the model predicts one of 31 transformation classes; in the generalization case, it predicts one of 10. For both cases, we average over a test set with size being (\# classes) times (\# original images), i.e., (\# classes) $\times$ 50k for ImageNet-1k. We measure semantic accuracy with ImageNet-1k class labels, separating accuracy on clean vs.~transformed (a.k.a., obfuscated) images.

\textbf{Embeddings}. 
While it is hard to control for all aspects of the foundation models, we enable a fair analysis for three image-only embeddings. We train the CAN, MAE, and SimCLR algorithms all on JFT-300M~\citep{sun_revisiting_2017}; they output a 1024-dim.~embedding from a ViT L/16. Compared to CLIP/ALIGN, the number of training images is also similar (300M vs 400M).
The SimCLR model also contains a projection to a 128-dim.~embedding that we use for one comparison. CLIP uses ViT L/14 for a 768-dim.~image embedding. ALIGN uses EfficientNet-L2 for the image encoder and outputs a 1376-dim.~embedding. 
We were given access to the ALIGN weights. Our baseline is a 1024-dim.~embedding from a supervised ViT L/16 trained on ImageNet-1k.

\textbf{Post-processing.} We pre-compute embeddings for all images and then ignore the pixels. We train a linear probe or a small MLP network on the frozen embeddings. Unless stated otherwise, the MLP has one hidden layer of width 2048, and we optimize with ADAM and 0.2 dropout. In the discussion below, we may use the embedding name interchangeably with the probe using 
that embedding. 


\begin{table}[t]
\caption{Transformation prediction accuracies for six embeddings on our transformed version of ImageNet-1k. Fine-grained has 31 classes (30 transforms), and generalization has 10 classes (28 training and 15 held-out test transforms). MLP, one hidden layer, width 2048. The right columns also present transformation accuracies on the subset of test data with unseen sub-transformations for two categories of the generalization task, noise and style transfer.
}
\label{table:transformations}
\vskip 0.15in
\begin{center}
\begin{small}
\begin{sc}
\begin{tabular}{l|cc|cc}
\toprule
Embedding & Fine-Grained & Generalization & Held-Out Noise & Held-Out Style \\
\midrule
CAN & \textbf{98.27} & 88.12 & 86.33 & 49.29 \\
MAE & 97.67 & 86.79 &  \textbf{94.67} & 28.92 \\
SimCLR & 93.05 & 87.32 & 59.55 & 54.27 \\
CLIP & 96.45 & \textbf{90.99} & 76.58 & \textbf{86.24}  \\
ALIGN & 96.66 & 89.22 & 85.69 & 69.61 \\
Supervised & 94.12 & 79.11 & 61.06  & 41.90 \\
\bottomrule
\end{tabular}
\end{sc}
\end{small}
\end{center}
\vskip -0.15in
\end{table}


\subsection{Do embeddings capture transformations? }

We compare the transformation prediction performance of six embeddings in \cref{table:transformations}. All embeddings perform extremely well on the transformation detection task: over 93\% accuracy for the fine-grained task and over 79\% accuracy for the generalization task. These embeddings preserve fairly detailed information about the input image that can be extracted with minimal post-processing (2-layer MLP).

\subsection{What is the most equivariant embedding?}

In the \textbf{fine-grained task} (a test of which embedding has the most detailed information about the image), the CAN probe performs the best with MAE being a close second. Note that both CAN and MAE use masking as part of the self-supervised training. This suggests that filling in the image patches increases the transformation sensitivity. The SimCLR embedding performs fairly well, despite the expectation that a contrastive loss would lead to high levels of invariance (we discuss SimCLR more in \cref{sec:exp:simclr}). CLIP and ALIGN perform slightly worse than CAN on the fine-grained task but still quite well, implying they are sensitive to 30 transformations.

In the \textbf{generalization task} (a test of how well an embedding's information about transformations can generalize), the two text-image embeddings (CLIP and ALIGN) perform better than all other methods. This suggests that training with text improves the generalization ability of the image embedding. We note that, for all methods, the decreased accuracy between fine-grained and generalization occurs because the held-out sub-transformations present a challenging OOD task. Also, all self-supervised models perform significantly better than the supervised baseline, suggesting that optimizing an embedding directly for semantic information does not by default retain as much transformation information. In \cref{sec:exp:mistakes} we analyze in detail the mistakes that the embeddings make.

\subsection{Isn't SimCLR supposed to be invariant?}
\label{sec:exp:simclr}
\cref{table:simclr} shows transformation prediction results for SimCLR. We consider two layers of the embedding model. Specifically, `SimCLR embed' refers to the second-to-last layer and has 1024 dimensions (which is standard and used in \cref{table:transformations}). Then, the network projects this onto the last layer `SimCLR proj' to form a 128 dim.~vector. We see `SimCLR embed' generally outperform `SimCLR proj' on both fine-grained and generalization, regardless of the post-processing method. We learn that the projection layer of SimCLR is responsible for much of the invariance that we expect from a contrastive loss. On the other hand, the layer right before this retains more information about transformations. 

We also control for the dimensionalities (1k vs.~128) by evaluating a network that has one hidden layer of width 128. With a small width, we still see a large improvement from using SimCLR embed vs.~proj (+14.88\% for fine-grained, +7.19\% for generalization). Finally, we can greatly improve the performance of SimCLR proj by post-processing with a width 16k network (+10.66\% for fine-grained, +5.81\% for generalization). This means that after the projection, there are transformation details that are not available via a linear probe but can be extracted with a 2-layer network.

\begin{table}[t]
\caption{SimCLR transformation prediction accuracies comparing the $1k$-dim embedding layer vs.~the  $128$-dim projection layer. Networks either have one hidden layer (width $\in\{128, 16k\}$) or a linear probe. Averaged over 5 runs, with all std.~dev.~below 0.14.}
\label{table:simclr}
\begin{center}
\begin{small}
\begin{sc}
\begin{tabular}{llcc}
\toprule
Embedding & Dim $\rightarrow$ Width & Fine-Grained & Generalization \\
\midrule
SimCLR embed & 1k $\rightarrow$ 16k & \textbf{93.55} & 86.97 \\
SimCLR embed & 1k $\rightarrow$ 128 & 90.67 & \textbf{87.61} \\
SimCLR embed & lin prob & 88.23 & 86.81\\
SimCLR proj & 128 $\rightarrow$ 16k & 82.83 & 82.87 \\
SimCLR proj & 128 $\rightarrow$ 128 & 75.79 & 80.42 \\
SimCLR proj & lin prob & 72.17 & 77.06 \\
\bottomrule
\end{tabular}
\end{sc}
\end{small}
\end{center}
\vskip -0.1in
\end{table}

\begin{table*}[ht]
\caption{Interplay between semantic (\textsc{Sem}), obfuscated semantic (\textsc{ObfSem}), and transformation (\textsc{Transf}) prediction accuracies on ImageNet-1k. \textsc{ObfSem} measures top-1 semantic accuracy on transformed images. We compare a 1-head model (semantic class only, one hidden layer, width 2048) to a 2-head model (multi-task semantic \& transformation, shared hidden layer, width $2048$). Entries `---' mean  network does not predict this. Averaged 5 runs, all semantic std.~dev.~below 0.13 and transform.~below 0.31. Best in each column in bold.}
\label{table:semantic}
\vskip 0.1in
\begin{center}
\begin{small}
\begin{sc}
\begin{tabular}{lcccc@{\hskip 0.3in}ccc}
\toprule
& \multicolumn{3}{c}{\bfseries Fine-Grained} && \multicolumn{3}{c}{\bfseries Generalization}\\
\bfseries Embedding & \bfseries Sem & \bfseries ObfSem & \bfseries Transf && \bfseries Sem & \bfseries ObfSem & \bfseries Transf \\
\midrule
CAN (sem only) & 75.95 & 66.25 & --- && 76.19 & 55.94 & --- \\
CAN (2-head) & 76.04 & 66.14 & \textbf{98.08} && 75.84 & 55.76 & 88.46 \\
\midrule
MAE (sem only) & 70.12 & 56.32 & --- && 69.85 & 44.17 & --- \\
MAE (2-head) & 70.18 & 56.08 & 97.35 && 69.99 & 44.31 & 87.90 \\
\midrule
SimCLR (sem only) & 74.52 & 63.31 & --- && 74.24 & 52.88 & --- \\
SimCLR (2-head) & 74.49 & 63.03 & 92.35 && 74.13 & 53.01 & 87.16 \\
\midrule
CLIP (sem only) & \textbf{84.56} & \textbf{71.57} & --- && \textbf{84.33} & \textbf{61.78} & --- \\
CLIP (2-head) & \textbf{84.43} & \textbf{71.38} & 95.51 && \textbf{84.28} & \textbf{61.73} & \textbf{90.76} \\
\midrule
ALIGN (sem only) & 84.24 & 70.35 & --- && 83.90 & 59.19 & --- \\
ALIGN (2-head) & 84.26 & 70.27 & 95.93 && 83.84 & 59.22 & 88.87 \\
\midrule
Supervised (sem only) & 77.80 & 62.87 & --- && 77.50 & 53.77 & --- \\
Supervised (2-head) & 77.69 & 62.67 & 93.07 && 77.47 & 53.77 & 78.66 \\
\bottomrule
\end{tabular}
\end{sc}
\end{small}
\end{center}
\end{table*}

\subsection{Does transformation information interfere with semantic information?}

We next explore the interplay between semantic and transformation accuracy by training two-head networks in a multi-task setting. The first head predicts the ImageNet-1k class. The second head predicts the transformation label. Both heads share the same 2048-dimensional hidden layer of the MLP that post-processes the embedding. As a baseline, we also train a one-head model that only predicts the semantic class (also using a 2-layer MLP, width 2048). We determine how the multi-task setting affects three metrics: semantic accuracy on clean images, obfuscated semantic accuracy on transformed images, and transformation prediction accuracy. \cref{table:semantic} reports these accuracies for the fine-grained and generalization versions. As before, we fix the embedding and only train the MLP.

\textbf{Clean semantic accuracy.} Using the two-head network leads to comparable semantic prediction compared to a one-head network. Only in some cases do we see a decrease in accuracy. The post-processing is able to effectively combine both semantic and transformation information in the MLP's hidden layer. Comparing semantic accuracies, CLIP and ALIGN outperform the other methods by a large margin. This is expected since the linear probe accuracy of vision-only self-supervised methods (CAN, MAE, SimCLR) tend to be lower than the accuracies after fine-tuning~\citep{mishra2022simple}. In \cref{sec:semantic-1k}, we show that a width $1k$ MLP leads to similar trends and accuracies.

\textbf{Obfuscated semantic accuracy.} We move on to discuss the semantic accuracy on the transformed images (\textsc{ObfSem}). In essence this is a metric for the robustness of the models to a dataset shift. Moreover, many of the transformations were not seen during training, and hence, we can consider the images to be OOD. Across the embeddings, we observe a mix of increases and decreases to the \textsc{ObfSem} accuracy. In general, the deviations are small, and we conclude that transformations sensitivity does not impact the ability to succeed at object-level predictions.

\subsection{Do all embeddings ignore the same transformation information?}
\label{sec:exp:mistakes}

We dig into the confusion matrices and how error trends reveal the information in embeddings. The fine-grained and generalization datasets lead to different insights, and we discuss them separately. 

\textbf{Fine-grained errors.} The most common mistakes for all embeddings come from (i) misclassifying medium Gaussian blur as low Gaussian blur, and (ii) underpredicting `Identity' for the unaltered images. Both mistakes are fairly expected. Comparing MAE to CAN, we find that MAE has worse performance for central cropping, which is likely due to its more aggressive masking during training (CAN uses 50\% masking while MAE uses 75\%). Considering SimCLR, the lower accuracy comes mostly from mispredicting hue shift, brighten, and saturate. For example, SimCLR labels 45\% images as `Identity' when their hue has been offset by~64. On the other hand, SimCLR performs comparably on the other transformations, including Gaussian blurring, despite this augmentation being part of the contrastive training. 
Compared to CAN and MAE, both CLIP and ALIGN have trouble with motion blur, perhaps because this is not an effect that is easily tied to textual cues.

\textbf{Generalization errors.} For two of the {held-out sub-transformations} (right two columns of \cref{table:transformations}), we see a large gap between the models. For the `noise' and `style transfer' generalization categories, we report the average per-class accuracy (for all images transformed in this way, the fraction that the model correctly predicts). Here, the chance level is $1/10$ since the model predicts among 10 categories. MAE performs the best for predicting the `noise' label, while CLIP is the best for the `style transfer' label. Hence the, MAE (resp.~CLIP) embedding encodes noise (resp.~style) information in a way that generalizes better to unseen examples of the same transformation category. 

\textbf{More about probing style transfer information.} 
CLIP performs quite well on the style transfer category, whereas this accounts for a sizeable fraction of errors for CAN, MAE, and Supervised.
For the held-out styles, CLIP correctly labels 86\% of images. The best vision-only model is SimCLR, which has 54\% accuracy. 
Hence, the CLIP embedding must not only capture the style, but do so in a way that the probe can recognize new styles from the features. This is not true for CAN/MAE, where 
the probe often predicts restyled images as unmodified (`Identity') or filtered (e.g, blurred). CLIP and SimCLR achieve over 70\% accuracy on the `Pasta' style, while CAN and MAE are below 4\%. \cref{sec:confusion-matrices} contains confusion matrices, and \cref{table:heldout} and \cref{table:style_transfer}, which evaluate held-out transformations.

\subsection{What have we learned about embeddings?}

\textbf{Robustness to domain shifts is not due to invariance.} By analyzing transformation prediction, we have drawn conclusions about the sensitivity of several embeddings along many axes. \cref{fig:summary} has summarized these insights, which help inform a choice between competing models. All of the models capture a lot of transformation information, which is quite unexpected given their robustness to OOD data like ImageNet variants. Sensitivity does \emph{not} imply poor generalization on OOD data.

\textbf{Modern embeddings suffice to detect OOD transformed data.} 
Embeddings can be used for classification of many transformations. 
Many classifiers are susceptible to transformation-based attacks, such as style transfer, Gaussian noise, or recoloring~\citep{cao2022stylefool,goodman2019cloud,hao2021s, hosseini2017google, li2019adversarial, yuan_2019_stealthy}. This is an important direction for content safety, beyond OOD and anomaly detection~\citep{salehi2021unified}. Our work shows that we can approach these tasks using MLPs and embeddings instead of pixel-based classifiers.

\textbf{Possible to have semantic \& transformation accuracy.} From \cref{table:semantic}, we see that the multi-task training leads to good performance on both semantic and transformation prediction. In some cases, the sensitivity to transformations even improves the obfuscated semantic accuracy. Another observation is that we achieve do the post-processing via the low-cost training of a 2-layer MLP. It is possible to fine-tune representations while freezing the large embedding model.

\section{Related Work}

\label{relatedwork}

\textbf{Probing and transformation prediction.} 
Compared to vision, much more work uses probes to understand language embeddings~\citep{belinkov2021probing, conneau2018you, hewitt2019designing}. Probing also led to new insights into generative models for code~\citep{lopez2022ast, troshin2022probing} and graph representations~\citep{akhondzadeh2023probing}.
On the vision side, work on visual chirality shows that it is possible to train a model to detect whether an image has been horizontally flipped~\citep{lin2020visual}. 
A related effort considers predicting domains like painting, sketch, or cartoon~\citep{zhu2022ood}. Researchers identify nuisance factors of X-ray images~\citep{sun2022beyond} with a pre-trained radiography model~\citep{sellergren2022simplified}. Training diffusion models involves reversing the (artificial) Gaussian noise in an image and optimization with a noise-prediction loss~\citep{ho2021classifier, ramesh2022hierarchical, song2019generative, yang2022diffusion}. Recent work on cold diffusion considers reversing other transformations, including deblurring, inpainting, super-resolution, and snow removal~\citep{bansal2022cold}. Building on previous work, we use transformation prediction to probe image embeddings with a broader set of transformations.
We also go beyond the type of content that can be probed by text prompts. 

\textbf{Foundation Models.} 
SimCLR~\citep{chen2020simple} trains on pairs of transformed images, penalizing the representation for differing embeddings. The embedding should be less sensitive to these transformations (cropping, color distortion, Gaussian blur). MAE~\citep{he2022masked} trains on images with patch-wise masking and reconstructs missing pixels.
Since the MAE model has to fill in parts of the image, it is possible that masking-based embeddings are more aware of visual distortions and stylization.
CAN~\citep{mishra2022simple} combines contrastive learning, masking, and noise prediction. Image embeddings also come from multi-modal models, such as CLIP~\citep{radford2021learning} and ALIGN~\citep{jia2021scaling}. Both use a contrastive loss to form visual-language representations of image-text pairs.  
The good performance and widespread use of foundation models motivates a deeper understanding of what retain or ignore. 
Work has also investigated  fine-tuning~\citep{evci2022head2toe, kumar2022fine} and  dataset quality~\citep{nguyenquality}. 

\textbf{Invariance and Equivariance.} The popularity of contrastive losses has led researchers to question whether embeddings should be insensitive (a.k.a., invariant) or sensitive (a.k.a., equivariant) to transformations~\citep{dangovski2021equivariant,duboisimproving, tian2020makes, xiao2021should}. This extends research on rotation prediction~\citep{metzger2020evaluating}, a seminal task for representation learning~\citep{komodakis2018unsupervised}. There has been efforts to measure equivariance through feature correlation~\citep{lenc2015understanding} or disentanglement~\citep{ngweta2023simple}, and to examine embeddings by reconstructing images~\citep{mahendran2015understanding}. Augmentation-aware learning has been proposed to improve semantic accuracy~\citep{metzger2020evaluating}. Another direction shows that 
contrastively learned domain-sensitive features may help OOD generalization~\citep{shen2022connect}. Compared to prior work, our probing task is more comprehensive. We evaluate a broader set of transformations (e.g., overlays, style transfer, dithering, solarize) as well as investigating more variation within similar categories (e.g., hue, saturation, brightness, blur).

\section{Conclusion}
\label{conclusion}

We constructed and investigated a new probing task that allowed us to shed new light on image embeddings. We showed that popular models capture enough information to distinguish dozens of transformations. Our experiments uncovered ways in which SimCLR is more invariant than CAN and MAE, and the types of transformations that are captured by self-supervised vision models vs.~image-text models, such as CLIP and ALIGN. One of our main takeways was that image-text embeddings encode style information in a way that generalizes better to unseen styles than image-only embeddings. We demonstrated that it is possible to post-process embeddings and extract more transformation information than a linear probe. The findings from the transformation probing provide new insights into the capacity of image embedding methods. Our results can help guide the choice of foundation models for downstream uses, such as for text-to-image generation models~\citep{chang2023muse, wei2023elite}. We discuss future work in \cref{sec:future-work}.

\bibliography{references}
\bibliographystyle{icml2023}

\newpage
\appendix
\onecolumn


\section{What Can You Do Next?}
\label{sec:future-work}
Our work is motivated by improving foundation models for their uses beyond semantic classification. We list many open directions for future work inspired by our findings:
\begin{enumerate}
    \item A central question is to create a nearly lossless image embedding that is also easily adapted for many downstream tasks. Our work suggests that it should be possible to keep more low-level information in the representation without compromising semantic performance. We believe this is an important direction because some of the downstream tasks may require these low-level features.
    \item Our results also suggest that networks that can predict transformations do not perform any worse in terms of data shift robustness (obfuscated semantic accuracy). This suggests that robust training methods might benefit from incorporating equivariance strategically, instead of focusing on invariance. Or, in contrast, transformers may be inherently equivariant, and achieving invariance may require even more aggressive training methods.
    \item Text-to-image generative models depend heavily on their pre-trained image encoder~\citep{chang2023muse, wei2023elite}. Fine-tuning the image backbone with transformation prediction could help in synthesizing transformed images. On the other hand, the invariance of image embeddings could prohibit the ability to generate certain visual features.
    \item A different direction is extending the transformation prediction task to be more fine-grained. We could ask the network to predict the specific parameters or strength of one or more transformations. One option is to predict both the transformation and strength of ImageNet-C transformations~\citep{hendrycks_benchmarking_2018}. This should make the task more challenging, and thus, reveal larger quantitative gaps in the performance of various embeddings.
    \item Another extension could be to identify which part of an image has been altered. This could uncover further differences between embedding methods. For example, masking-based embeddings might struggle with this given that they are trained on heavily obscured images. Image-text models might perform well because language cues can refer to parts of images and relative positioning of objects.
    \item An alternative way to probe the visual side of image-text models is through text prompts that describe visual aspects. This has been studied for some attributes like color, shape, and material~\citep{liu2022things, paik-etal-2021-world, zhang2022visual}. For example, \citet{zhang2022visual} uses questions like \emph{``what is the color of a penguin?''} or \emph{``what is the size of an ant?''}
    to probe the image-text model. Interestingly, our results suggest that CLIP and ALIGN retain quite a bit of visual information in the embedding. Hence, errors for the text prompts may be due to image-text alignment or to the language side of the model itself. It would be interesting to compare transformation prediction performance to these prompts and see if there are trends in the performance of different models.
    \item Recent work considers probes to understand how transformers process information~\citep{belrose2023eliciting}. In our SimCLR experiments, we saw that the final projection layer is responsible for much of the invariance. This suggests that certain layers may have a larger impact than others, as transformation information flows through the transformer model. Future work could continue the study of this interesting phenomenon.
    \item Considering other modalities, our transformation prediction task can apply to text or audio. For example, words can be changed with synonyms, characters can be replaced with symbols or typos, or sentences can be reordered based on syntactic freedoms. Following our analysis, this approach could then draw conclusions about the predispositions of different language models. This would complement some of the existing language model probing work~\citep{belinkov2021probing, bender2021dangers,conneau2018you, li2022emergent}.
    \item From an application point of view, it would be interesting to use transformation prediction for a data cleaning or filtering task. Another application is detecting (adversarial) image manipulations. 
    It is possible to use an MLP trained on top of an embedding to find anomalous images, such as those that have been stylized or heavily edited.
\end{enumerate}

\subsection{Where could we have done more?}

One deficiency of our work is that we have not proposed ways to translate our observations into improvements on benchmarks. Transformation awareness could improve performance on downstream tasks beyond ImageNet. It would ideal to couple transformation prediction with a new algorithm and create a self-supervised method that outperforms CAN, MAE, SimCLR, CLIP, and ALIGN.

Another shortcoming is that we have not actually used our models to detect a dataset shift in real-world data. There are many settings where discovering image transformations is important, including content safety, detecting copyright evasion, and discovering manipulation. In this direction, we have shown that different types of semantically-trained embeddings can perform well on  detection tasks. Our generalization task also shows that training even a small MLP on top of an embedding can suffice to detect held-out transformations.

\section{Experimental Set-Up}

\subsection{Drawing conclusions about embeddings}
\label{sec:drawing-conclusions}

We can use the transformation prediction task to measure if an embedding model captures certain visual content. Consider a transformation $t$, where $t(x)$ denotes the transformed version of $x$. Assume we can train a post-processing network to predict that $\phi(t(x))$ is transformed and $\phi(x)$ is not. Then, we can conclude that $\phi$ must preserve enough information about the image so $t$ can be detected. That is, $\phi(x) \neq \phi(t(x))$. More interestingly, a network may succeed at predicting most transformations $t$ from a set $\mathcal{T}$ when they are applied to images in a dataset $\mathcal{X}$. Hence, the sets $A_{t, \phi} = \{\phi(t(x)) \mid x \in \mathcal{X}\}$ for $t \in \mathcal{T}$ are mostly disjoint. It is possible to use a sample from $A_{t, \phi}$ to determine $t$ with high accuracy. We also believe the transformation prediction task is a direct measure of equivariance, as opposed to $k$-NN results~\citep{dangovski2021equivariant, xiao_what_2020}. 


If the network cannot detect the transformation $t$, then we may conclude the opposite. The embedding $\phi$ does not preserve enough information. We can further qualify this based on the amount of post-processing required to extract this information. If $t$ is detectable after zero or one layers, then the information must be readily accessible in $\phi(t(x))$. Otherwise, if $t$ can be detected but only after numerous layers, then the information is still present but can only be recovered after combining several sources of information from $\phi(t(x))$. If no amount of post-processing suffices, then the embedding must truly be invariant, and $\phi(x) \approx \phi(t(x))$.

Given the above discussion, the fine-grained and generalization tasks yield complementary insights. The benefit of the fine-grained task is that we can investigate the precision of the embedding's information. Distinguishing a blur of radius three vs.~five should require more detailed information than distinguishing blurring vs.~brightening. 
In the generalization task, the network does not see some sub-transformations during training, which enables us to measure a type of generalization. For example, consider transformations $t$ and $t'$ from the same class (e.g., two different styles). In the best case, we only use $t$ during training, and the network recognizes that $\phi(t'(x))$ is similar to $\phi(t(x))$. It could be that the embeddings are close together or that $\phi$ encodes the style in some way. On the other hand, the network may fail to generalize, and predict $\phi(t'(x))$ and $\phi(t(x))$ differently. One conclusion is that $\phi$ is insensitive to $t'$. However, we show later that prediction accuracy is quite high for the fine-grained task. The generalization mistakes actually imply that $\phi$ captures both transformations but does so in a divergent way. 

Interestingly, in contrast to some work in NLP probing~\citep{belinkov2021probing}, we observe the same trends using a linear probe and a 2-layer MLP. It would be worthwhile to also consider control metrics and random feature embeddings to further understand whether the transformation information is readily available or not, similar to \citet{hewitt2019designing}.

\subsection{More experiment and training details}

We use JAX to implement the models and run the experiments. The learning rate for is $10^{-3}$, and use per-device batch size of 1024 with roughly 41.2 epochs for the fine-grained dataset and 127.9 for coarse-grained. For both datasets, the model trains while seeing a total of roughly 1.6B examples. We do not use any warm-up steps.
All models are optimized with ADAM and the learning rate decreases linearly. Given that we are only training 2-layer MLPs, the wall clock time for training is under a few hours using TPUs.

We use dropout with rate $0.2$ for all experiments except when comparing SimCLR in \cref{table:simclr}, which has a dropout rate of $0.0$ because we compare with a linear probe. We experimented with deeper/wider networks and other dropout rates, but this did not lead to much different results.

For 2-head models, we sum the losses for the semantic prediction and transformation prediction tasks. We use categorical cross entropy for both losses on the separate tasks. We do not weight the losses separately. We also randomly sample batches without controlling for the distribution of transformations in each batch. Hence, for the fine-grained task, we expect $1/31$ of the images to be clean, and $1/10$ to be clean for the coarse-grained task.

Embeddings computed consistent with external implementations for the various embedding models. We were given access to the ALIGN weights, we also use a standard CLIP checkpoint. We trained the supervised model from scratch, without optimizing the data augmentation strategy.

We use three pre-trained models (CAN, MAE, SimCLR) that were trained by the authors of the CAN paper~\citep{mishra2022simple} and shared with us (all trained on JFT-300M). Interestingly, we achieve slightly higher top-1 accuracy on ImageNet-1k with our 2-head multi-task MLPs compared to the linear probe results that they report. Specifically, our MLPs on top of CAN, MAE, SimCLR have 76.04, 70.18, 74.49 top-1 accuracy, respectively. Their linear probe results for CAN, MAE, SimCLR are 75.4, 64.1, 73.4, respectively. This improvement could either be due to (i) the extra layer in the MLP or (ii) training the MLP with both clean and transformed data. The MAE paper reports higher accuracy than ours, with 73.5 top-1 linear probe~\citep{he2022masked}.


\section{Dataset and Transformation Details}
\label{dataset-details}

For the images, we use the ImageNet-1k dataset. We transform the images using standard methods in OpenCV~\citep{opencv_library} and Pillow~\citep{pillow_library}, and a few pre-trained models listed below.

\subsection{High-level motivation for our choice of transformations.}
\label{sec:transformation-motivation}

We aim to determine whether embeddings from image foundation models can be used to train classifiers for non-semantic tasks. This is advantageous because in large ML systems, embeddings are often pre-computed and stored for as a way to compress and pre-process image datasets. Hence, using a small MLP on top of an embedding offers a light-weight way to automatically compute predicted signals about the images.

There are many non-semantic tasks that would fit into this framework. For example, for data cleaning, it is important to recognize poor image quality (e.g., JPEG artifacts, motion blur, cropping, etc). For content filtering and policy enforcement, it may be crucial to detect image manipulations (e.g., style transfer, text/icon overlays). In general, non-semantic image information is crucial for a myriad of tasks, such as determining if an image is a painting or photograph, if it has been taken during the day or night, if it is high-fidelity or grainy, or if it has been edited from the original.

When choosing the sets of transformations, we have tried to cover a range of visual effects. Noise affects individual pixels and blurring affects nearby regions. Overlays are independent of the image, while style transfer heavily depends on the content. The filtering and quantizing options focus on hue, saturation, or value separately. Some transformations are barely human-visible, and others are strikingly obvious. Of course, the space of all possible transformations is impossible cover fully, but we aim to probe many aspects of embeddings.

In the generalization task, we have also tried to set-up an experiment that reflects real-world usage. For example, with style transfer, we train with a subset of styles and ask the model to recognize examples of transformed images with unseen styles. For the other categories, we also believe that \emph{quantizing} captures a variety of related recoloring effects, and filtering covers many types of blur and noise. There is certainly room to expand and refine the taxonomy of transformations, and this is a nice direction for future work.

\subsection{Fine-grained transformations}

Below is the list of transformations in the fine-grained transformation set. For transformations which are parameterized, multiple sets of parameters may be used. In this case, different parameter sets are considered as different "classes" in the transformation prediction problem. The parameters are the same for training and for testing.

\begin{itemize}
    \item Identity
    \begin{itemize}
        \item No transformation, i.e. the original images.
        \item No parameter.
    \end{itemize}
    \item Hue Scaling \& Shift
    \begin{itemize}
        \item Scale and shift the hue channel in the hue-saturation-lightness (HSL) color space. $hue_{new} = (hue \times scale + \mathtt{offset}) \text{ mod } 360$.
        \item Parameter set 1: $\mathtt{scale} = -32$, $\mathtt{offset} = -4$
        \item Parameter set 2: $\mathtt{scale} = 1$, $\mathtt{offset} = 64$.
    \end{itemize}
    \item Saturate \& Desaturate
    \begin{itemize}
        \item Scale and shift the saturation channel in the HSL color space. $saturation_{new} = clip(saturation \times scale + \mathtt{offset}, 0, 255)$.
        \item Parameter set 1: $\mathtt{scale} = 5$, $\mathtt{offset} = -4$
        \item Parameter set 2: $\mathtt{scale} = 0.25$, $\mathtt{offset} = 32$
    \end{itemize}
    \item Brighten \& Darken
    \begin{itemize}
        \item Shift the lightness channel in the HSL color space. $lightness_{new} = clip(lightness + \mathtt{offset}, 0, 255)$.
        \item Parameter set 1: $\mathtt{offset} = 96$
        \item Parameter set 2: $\mathtt{offset}$ is uniformly sampled between $-128$ and $-64$.
    \end{itemize}
    \item Gaussian Noise
    \begin{itemize}
        \item Add a random noise to each pixel. The noise distribution is a Gaussian distribution with mean $0$ and standard deviation $\sigma$.
        \item Parameter set 1: $\sigma = 0.05$
        \item Parameter set 2: $\sigma = 0.15$
    \end{itemize}
    \item Gaussian Blur
    \begin{itemize}
        \item Blur an image by a Gaussian function with a given radius.
        \item Parameter set 1: The radius is uniformly sampled between $3$ and $5$.
        \item Parameter set 2: The radius is uniformly sampled between $7$ and $9$.
    \end{itemize}
    \item Motion Blur
    \begin{itemize}
        \item Simulate a motion of an image (as a 2D rectangle) along a random direction by a given length (in pixels).
        \item Parameter set 1: $\mathtt{length} = 5$
        \item Parameter set 2: $\mathtt{length} = 10$
    \end{itemize}
    \item Corner Crop
    \begin{itemize}
        \item Keep only the bottom-right quadrant of an image.
        \item No parameter.
    \end{itemize}
    \item Rotation
    \begin{itemize}
        \item Rotate an image counter-clockwise by a given degree.
        \item The degree is uniformly sampled between $90$ and $270$.
    \end{itemize}
    \item JPEG Compression
    \begin{itemize}
        \item Re-compress an image with a given JPEG quality.
        \item The quality is uniformly sampled between $10$ and $15$.
    \end{itemize}
    \item Floyd-Steinberg Dithering~\citep{floyd_adaptive_1976}
    \begin{itemize}
        \item Reduce the bit depth of an image by applying the Floyd-Steinberg dithering algorithm.
        \item The bit depth is set to $1$.
    \end{itemize}
    \item Posterize
    \begin{itemize}
        \item Reduce the bit depth of an image by quantizing each pixel value independently.
        \item The bit depth is set to $2$.
    \end{itemize}
    \item Pixelate
    \begin{itemize}
        \item Create a pixelation effect by downsampling an image with a factor and then upsampling to its originnal size.
        \item The downsampling factor is set to $0.15$.
    \end{itemize}
    \item Solarize
    \begin{itemize}
        \item Simulate photo solarization by inverting each pixel value above a threshold.
        \item The threshold is set to $192$.
    \end{itemize}
    \item Grayscale
    \begin{itemize}
        \item Change an image to a grayscale image.
        \item No parameter.
    \end{itemize}
    \item Vertical Line Shift
    \begin{itemize}
        \item Rotate each column by a given distance (wrapping around), with even columns rotating down and odd columns rotating up.
        \item The distance is set to $3$.
    \end{itemize}
    \item Grid Overlay
    \begin{itemize}
        \item Change the pixels on even rows and on even columns to a fixed color $RGB=(204, 255, 127)$.
        \item No parameters.
    \end{itemize}
    \item Line Overlay
    \begin{itemize}
        \item Paint horizontal lines on an image.
        \item Each line is $4$-pixel wide, and the distance between adjacent lines is $20$ pixels. The lines are painted dark red $RGB = (101, 0, 0)$.
    \end{itemize}
    \item Icon Overlay
    \begin{itemize}
        \item Paint a wall of `grinning face' icons on an image.
        \item The opacity (alpha channel) of the icons is set to $32$. The width ratio between image and icon is set to $10$.
    \end{itemize}
    \item Text Overlay
    \begin{itemize}
        \item Paint a wall of constant gibberish text on an image.
        \item The text is colored dark gray $RGB=(25, 25, 25)$.
    \end{itemize}
    \item Line Halftoning
    \begin{itemize}
        \item Apply a halftone process based on amplitude-modulated sinusoidal waves~\citep{ahmed2016amplitude}.
        \item The waves are drawn with lines of $1$-pixel width, and the maximum amplitude is $5$ pixels.
    \end{itemize}
    \item Style Transfer
    \begin{itemize}
        \item Apply the style transfer model~\citep{ghiasi_exploring_2017} with a given style image.
        \item Parameter set 1: The style image is Vincent van Gogh's The Starry Night.
        \item Parameter set 2: The style image is Gyula Derkovits's Taligás.
        \item Parameter set 3: The style image is a photo of a bonfire.
        \item Parameter set 4: The style image is a photo of pasta.
    \end{itemize}
\end{itemize}

\subsection{Coarse-grained transformations}

Below is the list of transformation categories and sub-transformations in the coarse-grained transformation set. The testing set of some categories may contain more sub-transformations or wider parameter ranges. For randomized parameters, we use $U(a, b)$ to denote a uniform sample between $a$ and $b$ (inclusive). The parameters are independently sampled once for each image.

\begin{itemize}
    \item Identity
    \begin{itemize}
        \item No transformation, i.e. the original images.
        \item No parameter.
    \end{itemize}
    \item Icon Overlay
    \begin{itemize}
        \item Paint a wall of icons on an image.
        \item Training parameters: For each image an icon is randomly chosen from 5 candidate icons. The opacity (alpha channel) of the icons is $U(64, 128)$. The width ratio between image and icon is $U(8, 12)$.
        \item Testing parameters: Additional 5 candidate icons (10 in total). The opacity (alpha channel) of the icons is $U(64, 144)$. The width ratio between image and icon is $U(5, 15)$.
    \end{itemize}
    \item Line Halftoning
    \begin{itemize}
        \item Apply a halftone process based on amplitude-modulated waves~\citep{ahmed2016amplitude}.
        \item Training parameters: For each image an waveform is randomly chosen from 2 candidate waveforms. The waves are drawn with lines of $U(1, 2)$-pixel width, and the maximum amplitude is $U(5, 7)$ pixels.
        \item Testing parameters: Additional 2 candidate waveforms (4 in total). The waves are drawn with lines of $U(1, 2)$-pixel width, and the maximum amplitude is $U(4, 7)$ pixels.
    \end{itemize}
    \item Filtering: Transformations making the image blurry or less clear.
    \begin{itemize}
        \item Gaussian Blur
        \begin{itemize}
            \item Blur an image by a Gaussian function with a given radius.
            \item Training parameter: The radius is $U(3, 6)$.
            \item Testing parameter: The radius is $U(2, 9)$.
        \end{itemize}
        \item Motion Blur
        \begin{itemize}
            \item Simulate a motion of an image (as a 2D rectangle) along a random direction by a given length (in pixels).
            \item Training parameter: The radius is $U(18, 27)$.
            \item Testing parameter: The radius is $U(15, 35)$.
        \end{itemize}
        \item Pixelate
        \begin{itemize}
            \item Create a pixelation effect by downsampling an image with a factor and then upsampling to its originnal size.
            \item Training parameter: The downsampling factor is $U(0.25, 0.5)$.
            \item Testing parameter: The downsampling factor is $U(0.125, 0.5)$.
        \end{itemize}
        \item Blurry Background
        \begin{itemize}
            \item Change the aspect ratio and use a Gaussian blurred copy of the same image as background.
            \item Training parameters: Width and height scaling factors are $U(1.0, 1.8)$. Gaussian blur radius is $U(20, 40)$.
            \item Testing parameters: Width and height scaling factors are $U(0.7, 2.0)$. Gaussian blur radius is $U(20, 50)$.
        \end{itemize}
        \item Line Shift
        \begin{itemize}
            \item Rotate each row or column by a given distance (wrapping around), with even rows/columns rotating in an opposite direction to odd rows/columns.
            \item This transformation is held out in training.
            \item Testing parameter: The distance is $U(2, 8)$ pixels.
        \end{itemize}
    \end{itemize}
    \item Noise: Transformations adding high frequency artifacts.
    \begin{itemize}
        \item Gaussian Noise
        \begin{itemize}
            \item Add a random noise to each pixel. The noise distribution is a Gaussian distribution with mean $0$ and standard deviation $\sigma$.
            \item Training parameter: $\sigma \sim U(0.1, 0.5)$
            \item Testing parameter: $\sigma \sim U(0.1, 0.7)$
        \end{itemize}
        \item Impulse Noise  
        \begin{itemize}
            \item Randomly sample a percentage of pixels and paint half of them white and half of them black.
            \item Training parameter: Noise percentage is $U(10\%, 30\%)$.
            \item Testing parameter: Noise percentage is $U(5\%, 40\%)$.
        \end{itemize}
        \item Random Dithering  
        \begin{itemize}
            \item Quantize each pixel to $0$ or $255$ using a per-pixel random threshold.
            \item No parameters.
        \end{itemize}
        \item Ordered Dithering  
        \begin{itemize}
            \item Quantize each pixel to using a $2\times2$ Bayer threshold matrix~\citep{bayer1973optimum}.
            \item No parameters.
        \end{itemize}
        \item Floyd-Steinberg Dithering~\citep{floyd_adaptive_1976}
        \begin{itemize}
            \item Reduce the bit depth of an image by applying the Floyd-Steinberg dithering algorithm.
            \item This transformation is held out in training. 
            \item Testing parameter: The bit depth is $U(1, 2)$.
        \end{itemize}
    \end{itemize}
    \item Image Fusing: Transformations which fuse another image (as a distraction) in foreground or background.
    \begin{itemize}
        \item Image Overlay
        \begin{itemize}
            \item Add a small distraction image to foreground with partial opacity.
            \item Training parameters: 5 choices of distraction images. The distraction image's dimensions are $U(0.5, 0.7)$ fraction in size of the content image. The opacity is $U(64, 128)$.
            \item Testing parameters: Additional 6 choices of distraction images (11 in total). The distraction image's dimensions are $U(0.4, 0.8)$ fraction in size of the content image. The opacity is $U(64, 128)$.
        \end{itemize}
        \item Fusing
        \begin{itemize}
            \item Add a distraction image as background.
            \item Training parameters: 5 choices of distraction images. The foreground image's dimensions are $U(0.6, 0.8)$ fraction in size of the background image. The foreground's opacity is $U(128, 196)$.
            \item Testing parameters: Additional 6 choices of distraction images (11 in total). The foreground image's dimensions are $U(0.4, 0.9)$ fraction in size of the background image. The foreground's opacity is $U(128, 196)$.
        \end{itemize}
    \end{itemize}
    \item Quantizing: Transformations dealing with colors.
    \begin{itemize}
        \item Quantize Colors
        \begin{itemize}
            \item Reduce the number of distinct colors in an image. The colors are clustered and then replaced by the cluster centroids.
            \item Training parameter: The number of distinct colors after quantization is $U(16, 64)$.
            \item Testing parameters: The number of distinct colors after quantization is $U(8, 128)$.
        \end{itemize}
        \item Invert Colors
        \begin{itemize}
            \item Invert all pixel values.
            \item No parameter.
        \end{itemize}
        \item Solarize
        \begin{itemize}
            \item Simulate photo solarization by inverting each pixel value above a threshold.
            \item Training parameter: The threshold is $U(96, 192)$.
            \item Testing parameter: The threshold is $U(64, 224)$.
        \end{itemize}
        \item HSL To RGB
        \begin{itemize}
            \item Convert an image to the HSL color space, and then directly read the values as RGB.
            \item No parameter.
        \end{itemize}
        \item Grayscale
        \begin{itemize}
            \item Change an image to a grayscale image.
            \item This transformation is held-out in training.
            \item No parameter.
        \end{itemize}
        \item Hue Shift \& Scaling
        \begin{itemize}
            \item Scale and shift the hue channel in the hue-saturation-lightness (HSL) color space. $hue_{new} = (hue \times scale + \mathtt{offset}) \text{ mod } 360$.
            \item This transformation is held-out in training.
            \item Testing parameters 1: $\mathtt{scale}=1$, $\mathtt{offset}=U(60, 300)$.
            \item Testing parameters 2: $\mathtt{scale}=\pm U(8, 32)$, $\mathtt{offset}=U(0, 360)$.
        \end{itemize}
    \end{itemize}
    \item Static Overlay
    \begin{itemize}
        \item Line Overlay
        \begin{itemize}
            \item Paint a series of equidistant parallel lines on an image. The lines in one image are in a random direction and of the same random color.
            \item Training parameters: Each line is $U(5, 7)$-pixel wide, and the distance between adjacent lines is $U(18, 24)$ pixels.
            \item Testing parameters: Each line is $U(3, 10)$-pixel wide, and the distance between adjacent lines is $U(15, 30)$ pixels.
        \end{itemize}
        \item Text Overlay
        \begin{itemize}
            \item Paint a wall of gibberish text on an image. The text in one image are of the same random color.
            \item Training parameter: 5 choices of gibberish text.
            \item Testing parameter: Additional 5 choices of gibberish text (10 in total).
        \end{itemize}
        \item Grid Overlay
        \begin{itemize}
            \item Change the pixels on even rows and on even columns to a random color (sampled per image).
            \item No parameters.
        \end{itemize}
    \end{itemize}
    \item Style Transfer
    \begin{itemize}
        \item Arbitrary Neural Style Transfer
        \begin{itemize}
            \item Apply a style transfer model~\citep{ghiasi_exploring_2017} with a given style image.
            \item Training parameter 1: The style image is Vincent van Gogh's The Starry Night.
            \item Training parameter 2: The style image is Gyula Derkovits's Taligás.
            \item Training parameter 3: The style image is Edvard Munch's The Scream.
            \item Training parameter 4: The style image is Katsushika Hokusai's The Great Wave off Kanagawa.
            \item Held-out parameter 1: The style image is Amadeo de Souza-Cardoso's Landscape with Black Figure.
            \item Held-out parameter 2: The style image is Pablo Picasso's Violon.
            \item Held-out parameter 3: The style image is a photo of a a bonfire.
            \item Held-out parameter 4: The style image is a photo of pasta.
        \end{itemize}
        \item Artistic Style Transfer
        \begin{itemize}
            \item Apply a style transfer model~\citep{johnson_perceptual_2016} with a set of pre-trained weights.
            \begin{itemize}
                \item Training parameter 1: The weights are pre-trained to mimic stained glass mosaics.
                \item Training parameter 2: The weights are pre-trained toward Francis Picabia's Udnie.
                \item Held-out parameter 1: The weights are pre-trained toward Vincent van Gogh's The Starry Night.
                \item Held-out parameter 2: The weights are pre-trained toward a painting of candies.
            \end{itemize}
        \end{itemize}
        \item Deep Dream
        \begin{itemize}
            \item Run a pre-trained DeepDream model~\citep{alexander_inceptionism_2015} to enhance the patterns that the model recognizes.
            \item This transformation is held-out in training.
            \item Testing parameter: The DeepDream process is configured with $U(7, 12)$ update iterations, learning rate $U(0.05, 0.08)$, number of octaves $U(6, 12)$, and octave scale $U(1.5, 2.0)$.
        \end{itemize}
    \end{itemize}
    \item Warping: Transformations which rotate or transpose images.
    \begin{itemize}
        \item Rotation
        \begin{itemize}
            \item Rotate an image counter-clockwise by a given degree.
            \item Train parameter: The degree of rotation is $90$.
            \item Held-out parameter 1: The degree of rotation is $180$.
            \item Held-out parameter 2: The degree of rotation is $270$.
        \end{itemize}
        \item Vertical Flip
        \begin{itemize}
            \item Flip an image top to bottom.
            \item No parameter.
        \end{itemize}
        \item Transpose
        \begin{itemize}
            \item Flip an image diagonally along a diagonal.
            \item Training parameter: Flipping along the minor diagonal.
            \item Held-out parameter: Flipping along the major diagonal.
        \end{itemize}
    \end{itemize}
\end{itemize}




\newpage
\section{Further Experimental Results}
\label{sec:confusion-matrices}

\subsection{Smaller Representation (width 1k MLP)}
\label{sec:semantic-1k}
\cref{table:semantic-1k} shows the one-head and two-head accuracies for an MLP with one hidden layer of width 1k (the main paper table had width 2k). We use a one-headed model for each of transformation and semantic prediction, and then a two-headed model that trains to predict both labels. For the fine-grained dataset, performance is comparable for one- and two-head models. However, the two-headed models underperforms slightly (less than 1\% worse). Interestingly, the coarse-grained dataset has the opposite trend for CAN, MAE, and SimCLR when it comes to transformation accuracy. Here, the two-headed model actually leads to a significant improvement in transformation prediction accuracy. The multi-task set-up likely prevents overfitting. This is beneficial because the coarse-grained dataset has held-out transformations (whereas the fine-grained dataset has the same train/test transformations).

\begin{table*}[ht]
\caption{Reducing the width to 1024 to evaluate a smaller  post-processed representation for the interplay between semantic (\textsc{Sem}), obfuscated semantic (\textsc{ObfSem}), and transformation (\textsc{Transf}) prediction accuracies on ImageNet-1k. \textsc{ObfSem} measures top-1 semantic accuracy on transformed images. We compare a 1-head model (semantic class only, one hidden layer, width 1024) to a 2-head model (multi-task semantic \& transformation, shared hidden layer, width $1024$). Entries `---' mean  network does not predict this. Averaged 5 runs, all semantic std.~dev.~below 0.18 and transform.~below 0.27. Best in each column in bold and within one standard deviation.}
\label{table:semantic-1k}
\vskip 0.1in
\begin{center}
\begin{small}
\begin{sc}
\begin{tabular}{lcccc@{\hskip 0.3in}ccc}
\toprule
& \multicolumn{3}{c}{\bfseries Fine-Grained} && \multicolumn{3}{c}{\bfseries Generalization}\\
\bfseries Embedding & \bfseries Sem & \bfseries ObfSem & \bfseries Transf && \bfseries Sem & \bfseries ObfSem & \bfseries Transf \\
\midrule
CAN (tran only) &  --- & --- & \textbf{98.25} && --- & --- & 87.78 \\
CAN (sem only) & 75.89 & 65.33 & --- && 75.52 & 55.04 & --- \\
CAN (2-head) & 75.79 & 65.21 & 97.73 && 75.45 & 54.99 & 88.44 \\
\midrule
MAE (tran only) &  --- & --- & 97.53 && --- & --- & 86.60 \\
MAE (sem only) & 69.36 & 54.87 & --- && 68.60 & 42.93 & --- \\
MAE (2-head) & 68.84 & 54.36 & 97.01 && 68.55 & 42.83 & 87.22 \\
\midrule
SimCLR (tran only) &  --- & --- & 92.83 && --- & --- & 86.60 \\
SimCLR (sem only) & 74.20 & 62.33 & --- && 73.87 & 52.28 & --- \\
SimCLR (2-head) & 73.92 & 61.92 & 91.79 && 73.75 & 52.23 & 88.14 \\
\midrule
CLIP (tran only) &  --- & --- & 96.32 && --- & --- & \textbf{90.78} \\
CLIP (sem only) & \textbf{84.26} & \textbf{71.01} & --- && \textbf{84.00} & \textbf{61.23} & --- \\
CLIP (2-head) & \textbf{84.03} & \textbf{70.78} & 95.32 && \textbf{83.98} & \textbf{61.16} & \textbf{90.62} \\
\midrule
ALIGN (tran only) &  --- & --- & 96.44 && --- & --- & 89.04 \\
ALIGN (sem only) & \textbf{84.24} & 69.91 & --- && \textbf{83.95} & 59.05 & --- \\
ALIGN (2-head) & \textbf{84.24} & 69.78 & 95.61 && \textbf{83.92} & 59.09 & 88.96 \\
\midrule
Supervised (tran only) &  --- & --- & 93.86 && --- & --- & 78.93 \\
Supervised (sem only) & 77.87 & 62.39 & --- && 77.61 & 53.66 & --- \\
Supervised (2-head) & 77.76 & 62.07 & 92.70 && 77.49 & 53.57 & 78.51 \\
\bottomrule
\end{tabular}
\end{sc}
\end{small}
\end{center}
\vskip -0.1in
\end{table*}

\subsection{Analyzing the Held-Out Transformations and Generalization Dataset}
\begin{figure*}[ht]
\centering
\subfloat[Identity\label{fig:st:identity}]{\includegraphics[width=0.24\textwidth]{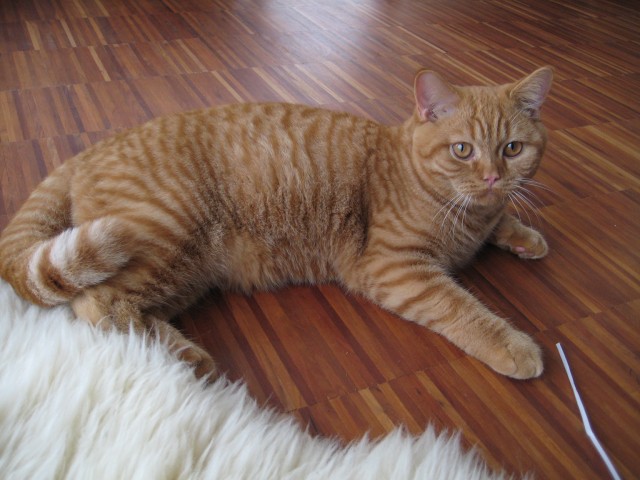}}\hfill
\subfloat[Artistic Mosaic\label{fig:st:mosaic}] {\includegraphics[width=0.24\textwidth]{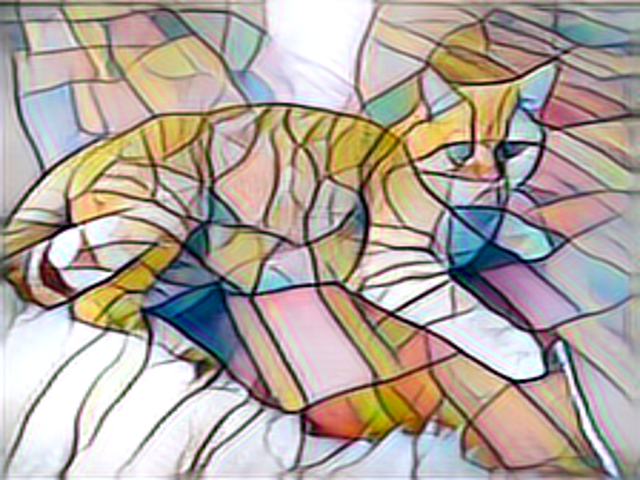}}\hfill
\subfloat[Artistic Udnie\label{fig:st:udnie}] {\includegraphics[width=0.24\textwidth]{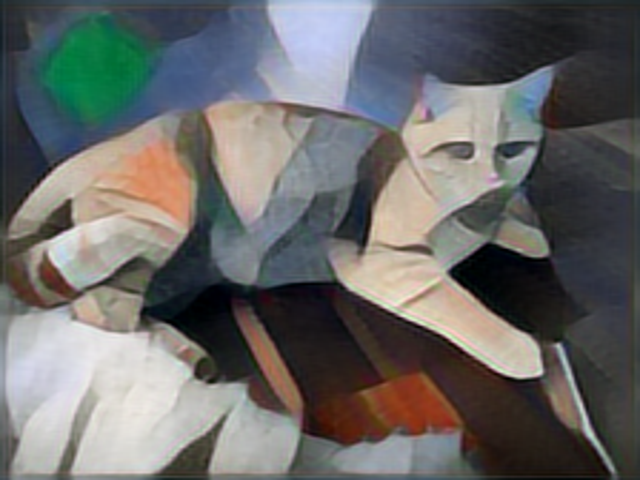}}\hfill
\subfloat[Great Wave Off Kanagawa\label{fig:st:greatwave}] {\includegraphics[width=0.24\textwidth]{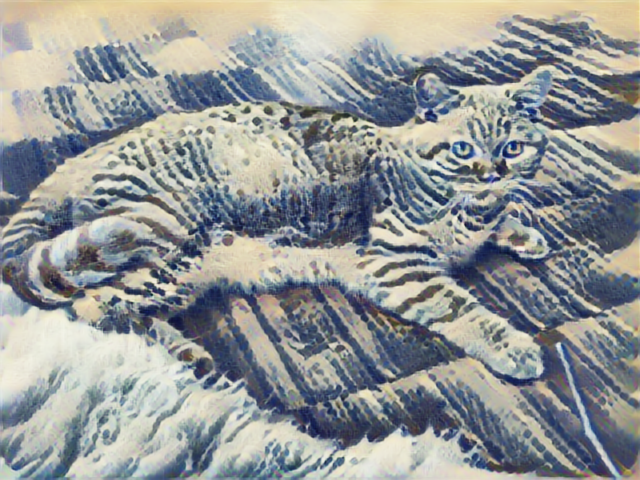}}\hfill
\subfloat[The Scream\label{fig:st:thescream}] {\includegraphics[width=0.24\textwidth]{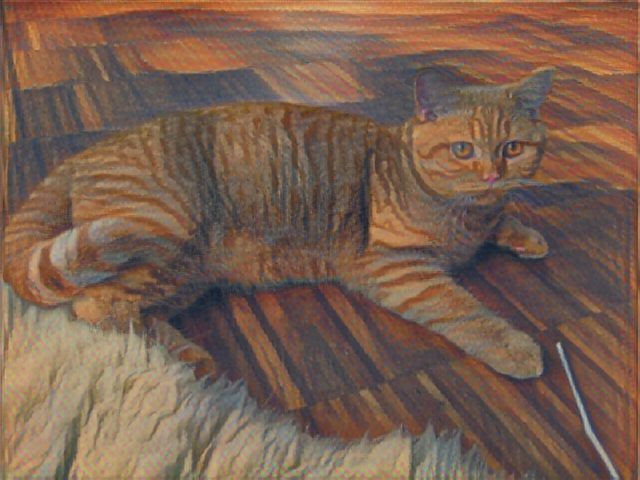}}\hspace{.0134\textwidth}
\subfloat[Starry Night\label{fig:st:starrynight}]{\includegraphics[width=0.24\textwidth]{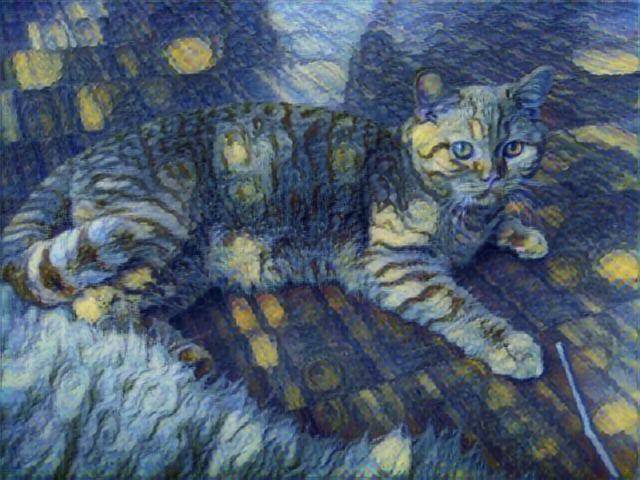}}\hspace{.0134\textwidth}
\subfloat[Taligas 1920\label{fig:st:taligas}] {\includegraphics[width=0.24\textwidth]{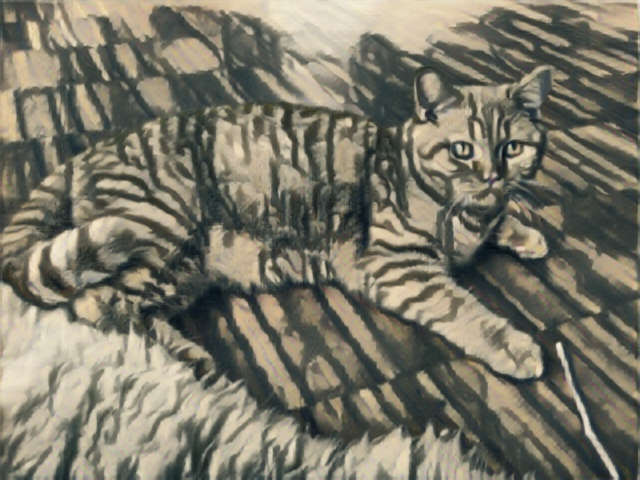}}\hfill

\vspace{0.5in}
\subfloat[Artistic Candy\label{fig:st:candy}] {\includegraphics[width=0.24\textwidth]{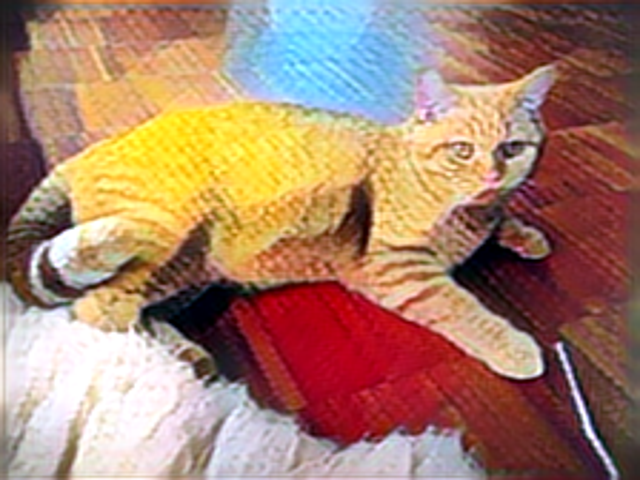}}\hfill
\subfloat[Artistic Starry Night\label{fig:st:artisticstarrynight}] {\includegraphics[width=0.24\textwidth]{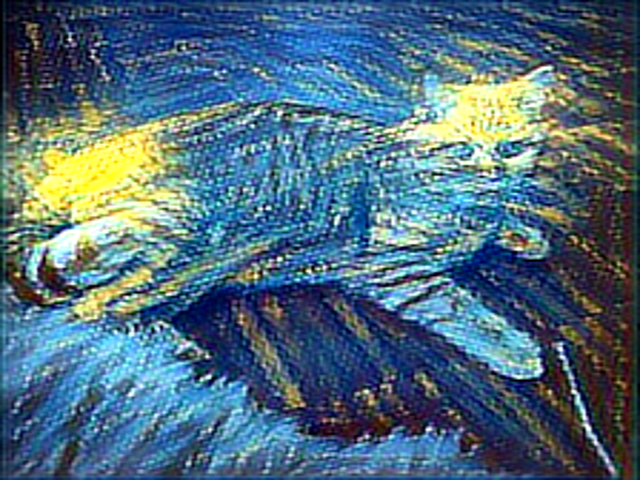}}\hfill
\subfloat[Bonfire\label{fig:st:bonfire}] {\includegraphics[width=0.24\textwidth]{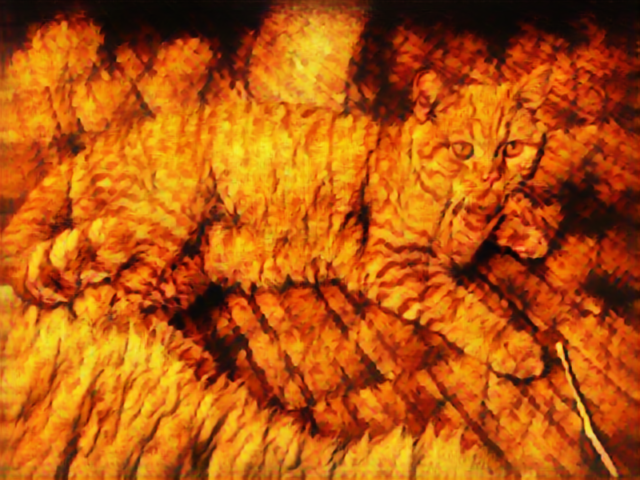}}\hfill
\subfloat[Deep Dream\label{fig:st:deepdream}] {\includegraphics[width=0.24\textwidth]{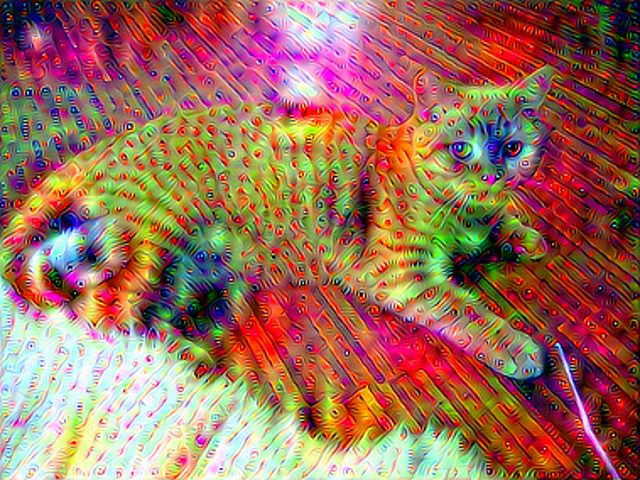}}\hfill
\subfloat[Landscape Black Figure\label{fig:st:blackfigure}] {\includegraphics[width=0.24\textwidth]{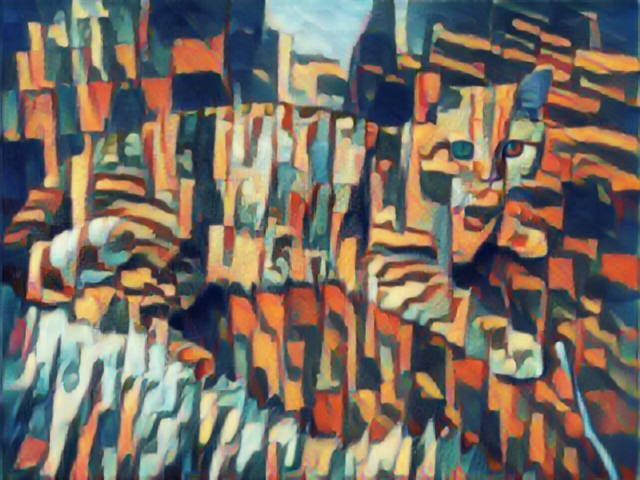}}\hspace{.0134\textwidth}
\subfloat[Pasta\label{fig:st:pasta}] {\includegraphics[width=0.24\textwidth]{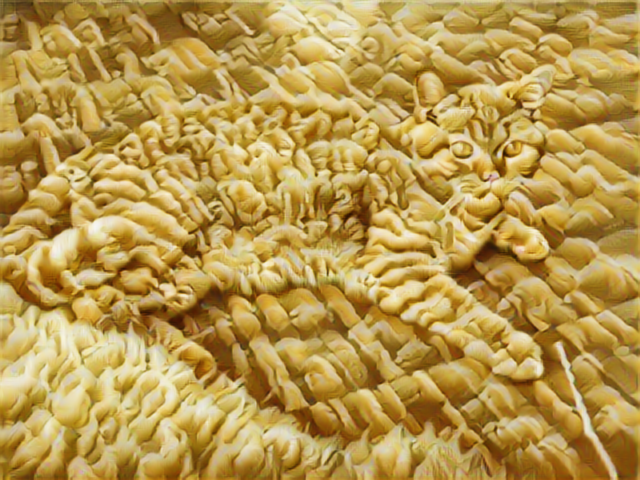}}\hspace{.0134\textwidth}
\subfloat[Violon\label{fig:st:violon}] {\includegraphics[width=0.24\textwidth]{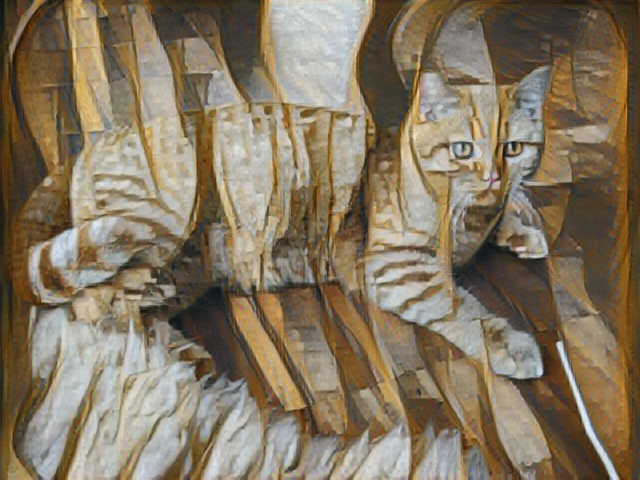}}\hfill
\caption{Style sub-transformations from the `Style Transfer' category of the coarse-grained dataset. We also include the `Identity' as the original image for reference. The six styles in the top group are in both the train and test sets, while the bottom seven styles only appear in the test set. All styles have the `Style Transfer' label. \cref{table:style_transfer} has accuracies for the different embeddings for each of these styles.
}
\label{fig:style_transfer}
\end{figure*}

\cref{table:heldout} presents average accuracies for the held-out sub-transformations. \cref{table:style_transfer} zooms in the on the style transfer accuracies, showing the fraction of correct prediction for each of the thirteen styles that are displayed in \cref{fig:style_transfer}. Then, we present the $10 \times 10$ confusion matrices for the coarse-grained dataset, one for each embedding. This matrices demonstrate the common errors made by the models, which inform the ways that the probe uncovers properties of the embeddings. Specifically, mispredicting certain transformations as `Identity' either points to invariance or to an inconsistent encoding of the transformation information.

\begin{table*}[ht]
\caption{Held-out transformation prediction accuracies for six embeddings on the coarse-grained dataset. For each transformation category, we report the average per-class accuracy (i.e., for all images that have been transformed in this way, we report the fraction that the model correctly predicts). We average over all held-out sub-transformations in each category. Interestingly, CLIP is best in two categories, while MAE, CAN, and Supervised are all best in one. Also, the relative performance of the methods has much larger gaps than the previous fine-grained and coarse-grained analyses. Image-text models perform much better for style transfer than the image-only models (see \cref{table:style_transfer} for individual style accuracies). We can see some fairly large differences between CLIP and ALIGN as well.}
\label{table:heldout}
\vskip 0.1in
\begin{center}
\begin{small}
\begin{sc}
\begin{tabular}{lccccc}
\toprule
Embedding & Filtering & Noise & Quantizing & Style Transfer & Warping \\
\midrule
CAN & 2.64 & 86.33 & 22.81 & 49.29 & \textbf{97.43} \\
MAE & 4.11 & \textbf{94.67} & 21.74 & 28.92 & 96.38 \\
SimCLR & 6.99 & 59.55 & 33.07 & 54.27 & 95.30 \\
CLIP & 9.83 & 76.58 & \textbf{51.78} & \textbf{86.24} & 93.94 \\
ALIGN & 3.24 & 85.69 & 30.28 & 69.61 & 92.60 \\
Supervised & \textbf{11.95} & 61.06 & 50.70 & 41.90 & 95.27 \\
\bottomrule
\end{tabular}
\end{sc}
\end{small}
\end{center}
\vskip -0.1in
\end{table*}


\begin{table*}[ht]
\caption{Accuracies for the style transfer sub-transformations in the coarse-grained dataset. For all images that have been transformed in this way, we report the fraction for which the model predicts `Style Transfer' correctly. The top six styles are in both the train and test sets, while the bottom seven styles only appear in the test set. 
When the styles are seen in the train set, then the validation accuracy is nearly perfect and often 100\% for these six styles. For the held-out styles, CLIP performs the best, sometimes by a large margin. Interestingly, SimCLR also performs well, often better than CAN, MAE, and Supervised. When the models mispredict style transfer as another label, the most common are `Identity' and `Filter', where `Filter' includes pixelation and various blurs.}
\label{table:style_transfer}
\vskip 0.1in
\begin{center}
\begin{small}
\begin{sc}
\begin{tabular}{lcccccc}
\toprule
Style & CAN & MAE & SimCLR & CLIP & ALIGN & Supervised \\
\midrule
Artistic Mosaic & \textbf{100} & \textbf{100}  & \textbf{100} & \textbf{100} & \textbf{100} & \textbf{100} \\
Artistic Udnie & \textbf{100} & \textbf{100}  & \textbf{100} & \textbf{100} & 99.99 & 99.99 \\
Great Wave Off Kanagawa & \textbf{100} & \textbf{100}  & \textbf{100} & \textbf{100} & \textbf{100} & \textbf{100} \\
The Scream & \textbf{100} & \textbf{100}  & 99.97 & 99.97 & 99.95 & 99.97 \\
Starry Night & 99.98 & \textbf{100}  & 99.98 & 99.98 & 99.98 & 99.97 \\
Taligas 1920 & 99.99 & \textbf{100}  & 99.99 & 99.99 & 99.99 & 99.99 \\
\midrule
Artistic Starry Night & 53.66 & 47.13 & 5.91 & \textbf{91.33} & 70.91 & 90.75 \\
Artistic Candy & 55.09 & 37.09 & 79.00 & \textbf{95.09} & 77.48 & 6.47 \\
Bonfire & 46.33 & 0.00 & 2.33 & \textbf{92.12} & 69.24 & 9.93 \\
Deep Dream & 3.13 & 0.12 & 26.34 & \textbf{52.10} & 11.84 & 3.40 \\
Landscape Black Figure & 99.92 & 98.94 & 99.75 & \textbf{99.99} & 99.97 & 79.43 \\
Pasta & 3.73 & 0.46 & 70.07 & \textbf{73.39} & 58.87 & 39.84 \\
Violon & 83.15 & 18.67 & 96.46 & \textbf{99.69} & 98.98 & 63.49 \\
\bottomrule
\end{tabular}
\end{sc}
\end{small}
\end{center}
\vskip -0.1in
\end{table*}

\begin{figure*}[ht]
\centering
\includegraphics[width=0.8\textwidth]{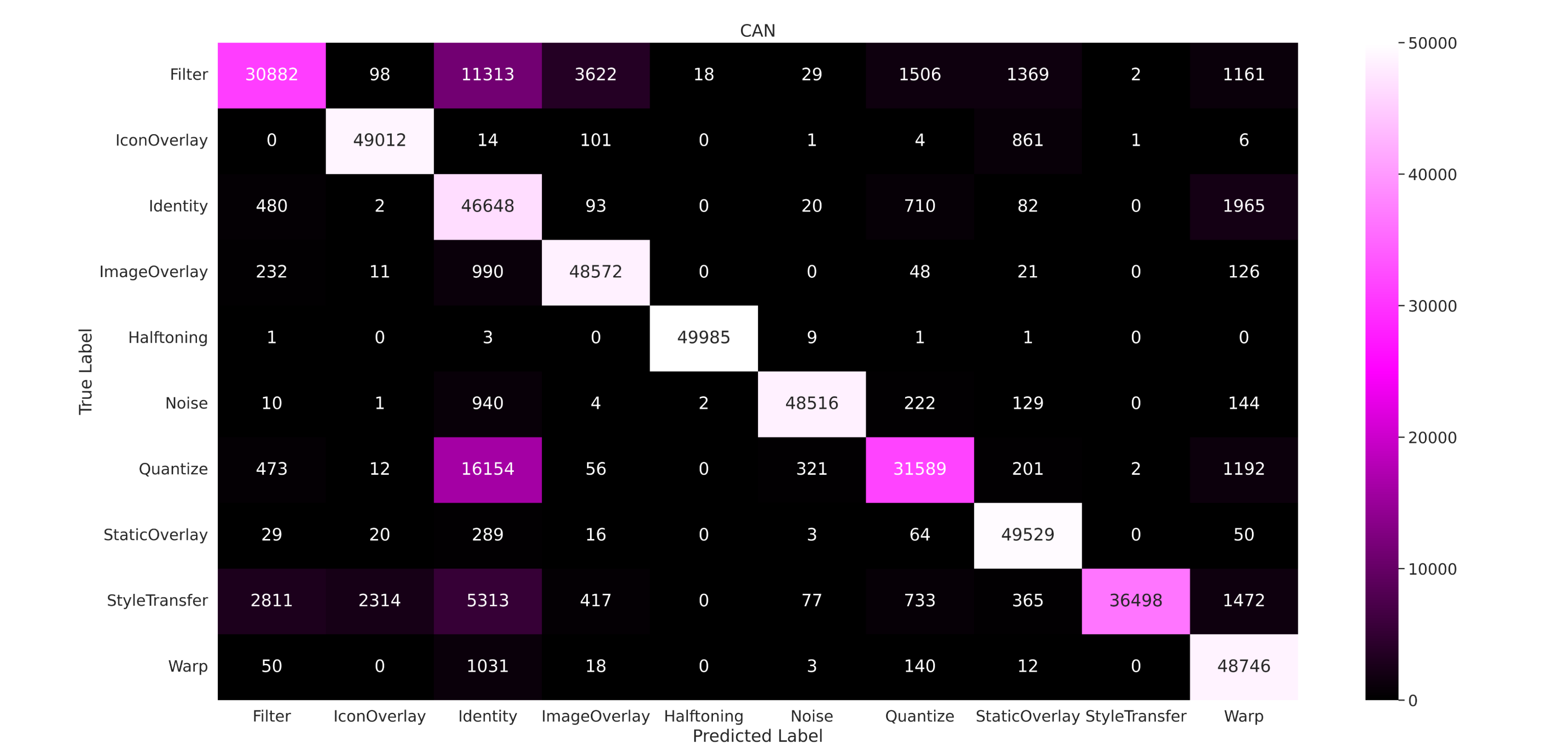}
\caption{CAN coarse-grained confusion matrix. Rows correspond to the 50k images of each of the 10 labels, and columns correspond to the number of predictions from the model for each of the 10 categories.}
\label{fig:can-confusion-coarse}
\vskip -0.2in
\end{figure*}

\begin{figure*}[ht]
\centering
\includegraphics[width=0.8\textwidth]{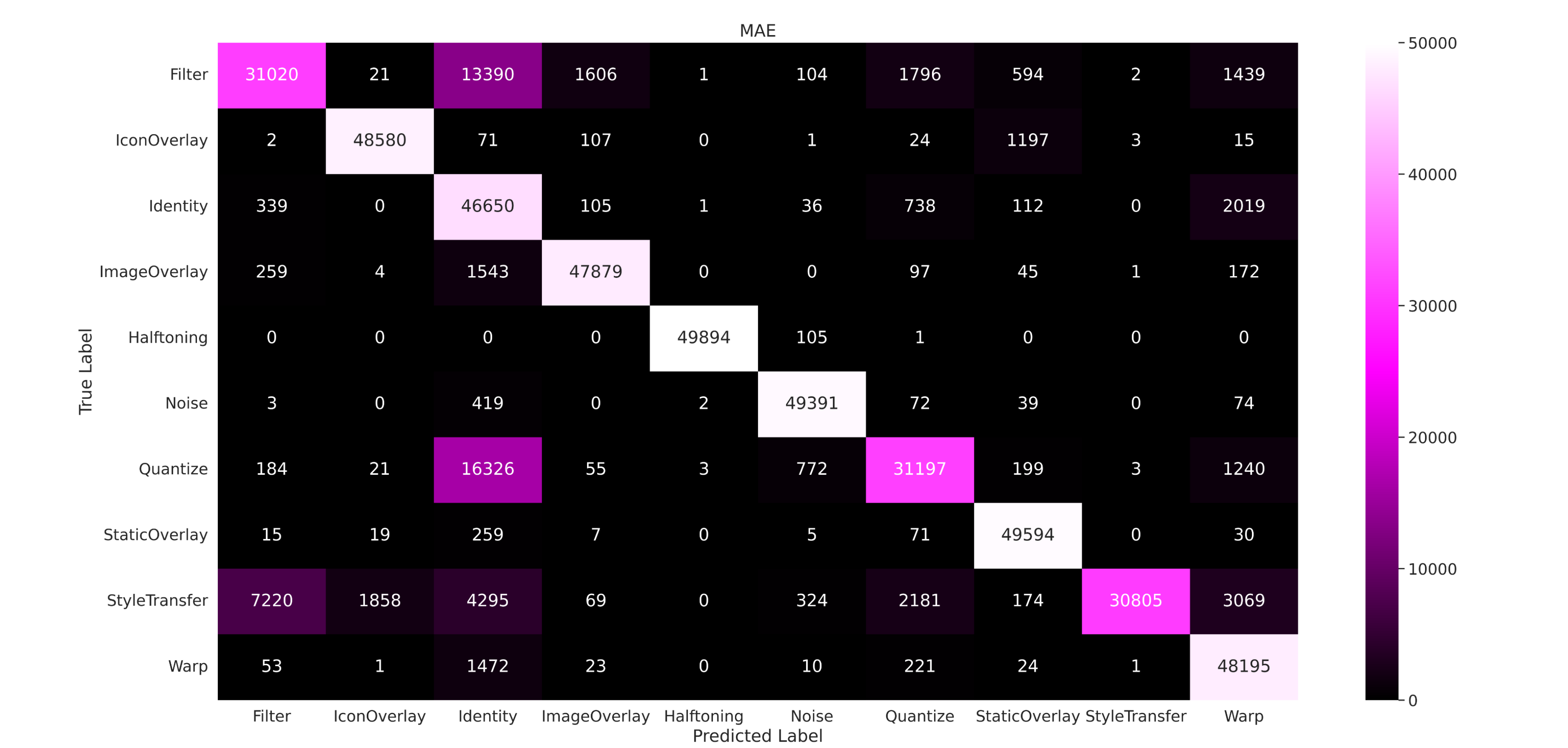}
\caption{MAE coarse-grained confusion matrix. Rows correspond to the 50k images of each of the 10 labels, and columns correspond to the number of predictions from the model for each of the 10 categories.}
\label{fig:MAE-confusion-coarse}
\vskip -0.2in
\end{figure*}

\begin{figure*}[ht]
\centering
\includegraphics[width=0.8\textwidth]{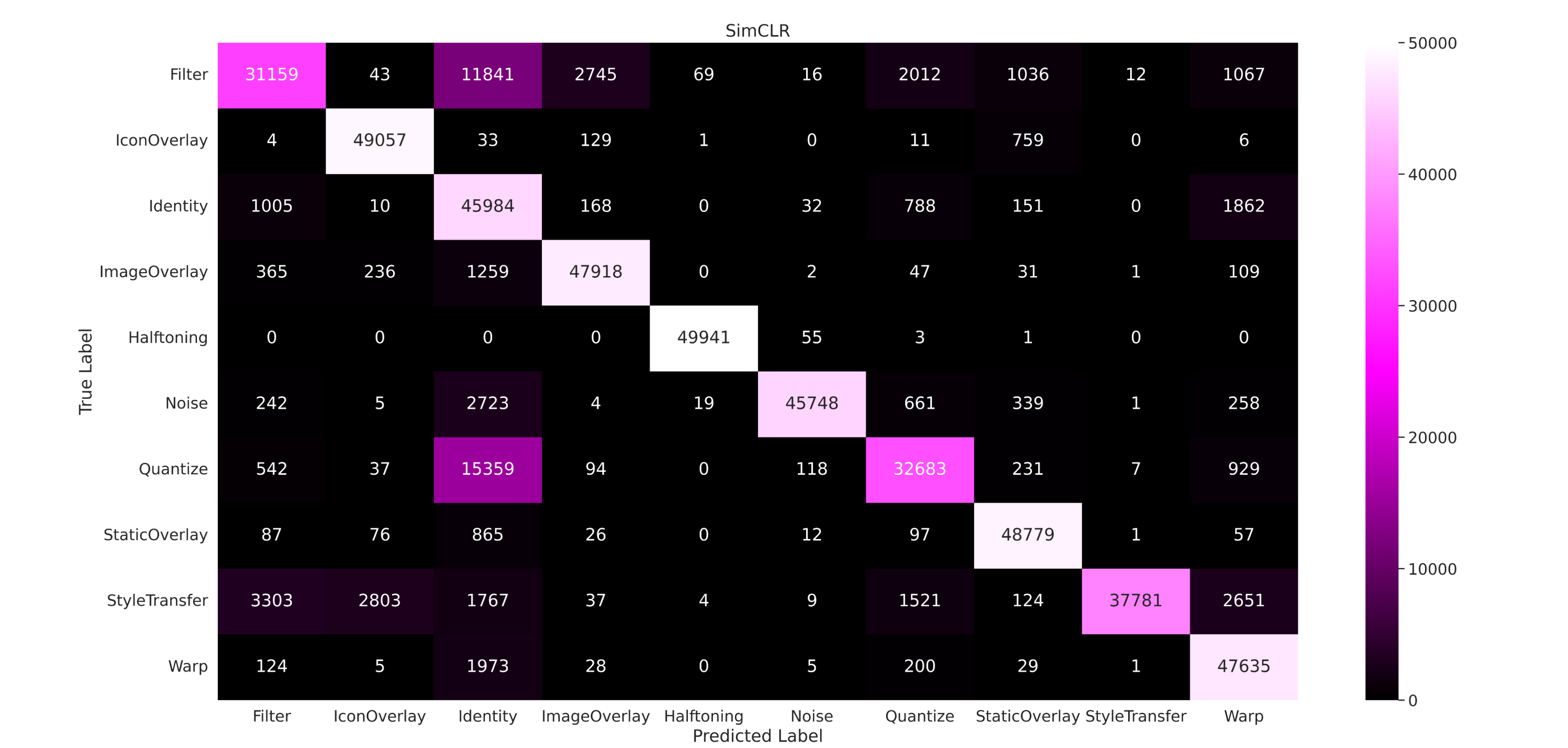}
\caption{SimCLR coarse-grained confusion matrix. Rows correspond to the 50k images of each of the 10 labels, and columns correspond to the number of predictions from the model for each of the 10 categories.}
\label{fig:simclr-confusion-coarse}
\vskip -0.2in
\end{figure*}

\begin{figure*}[ht]
\centering
\includegraphics[width=0.8\textwidth]{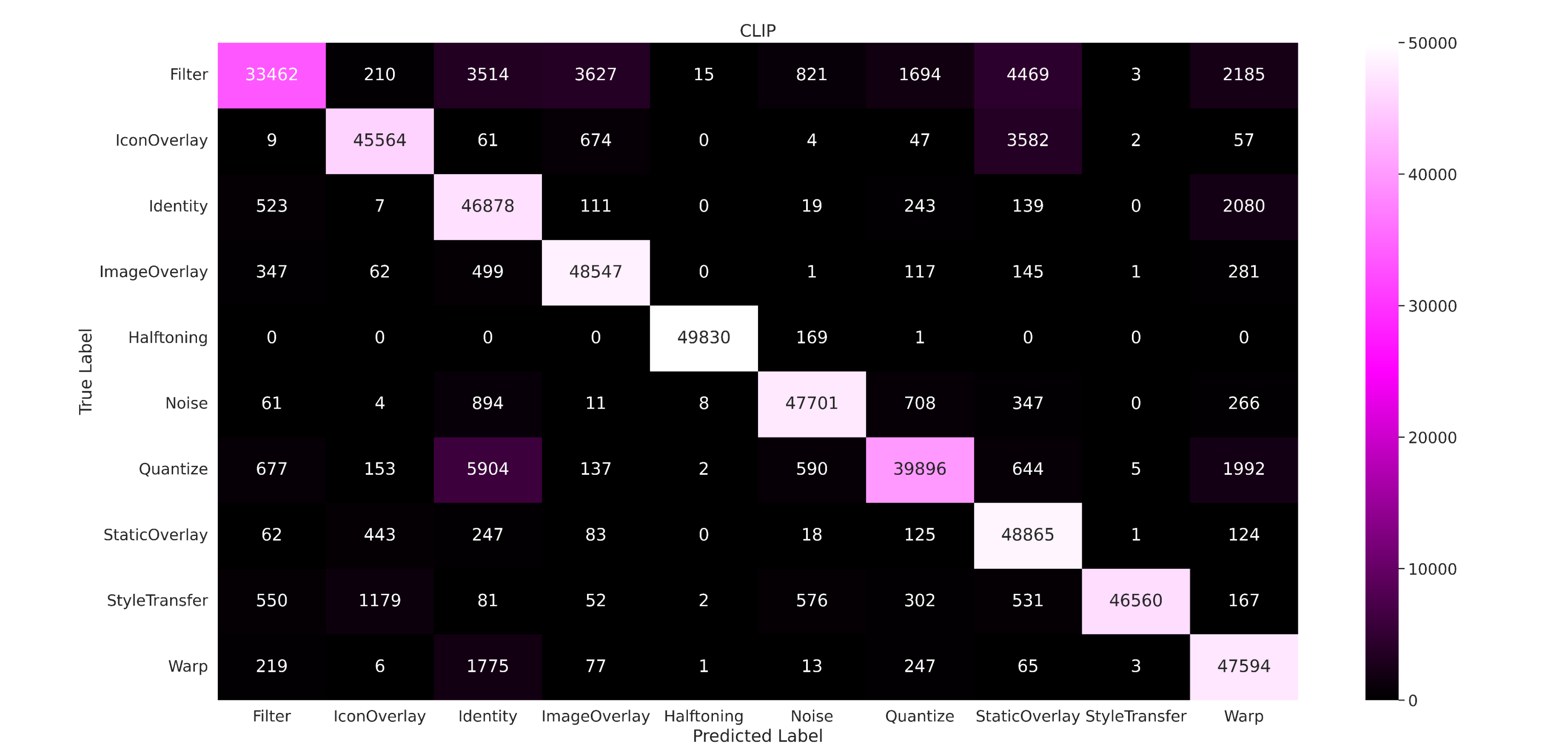}
\caption{CLIP coarse-grained confusion matrix. Rows correspond to the 50k images of each of the 10 labels, and columns correspond to the number of predictions from the model for each of the 10 categories.}
\label{fig:clip-confusion-coarse}
\vskip -0.2in
\end{figure*}

\begin{figure*}[ht]
\centering
\includegraphics[width=0.8\textwidth]{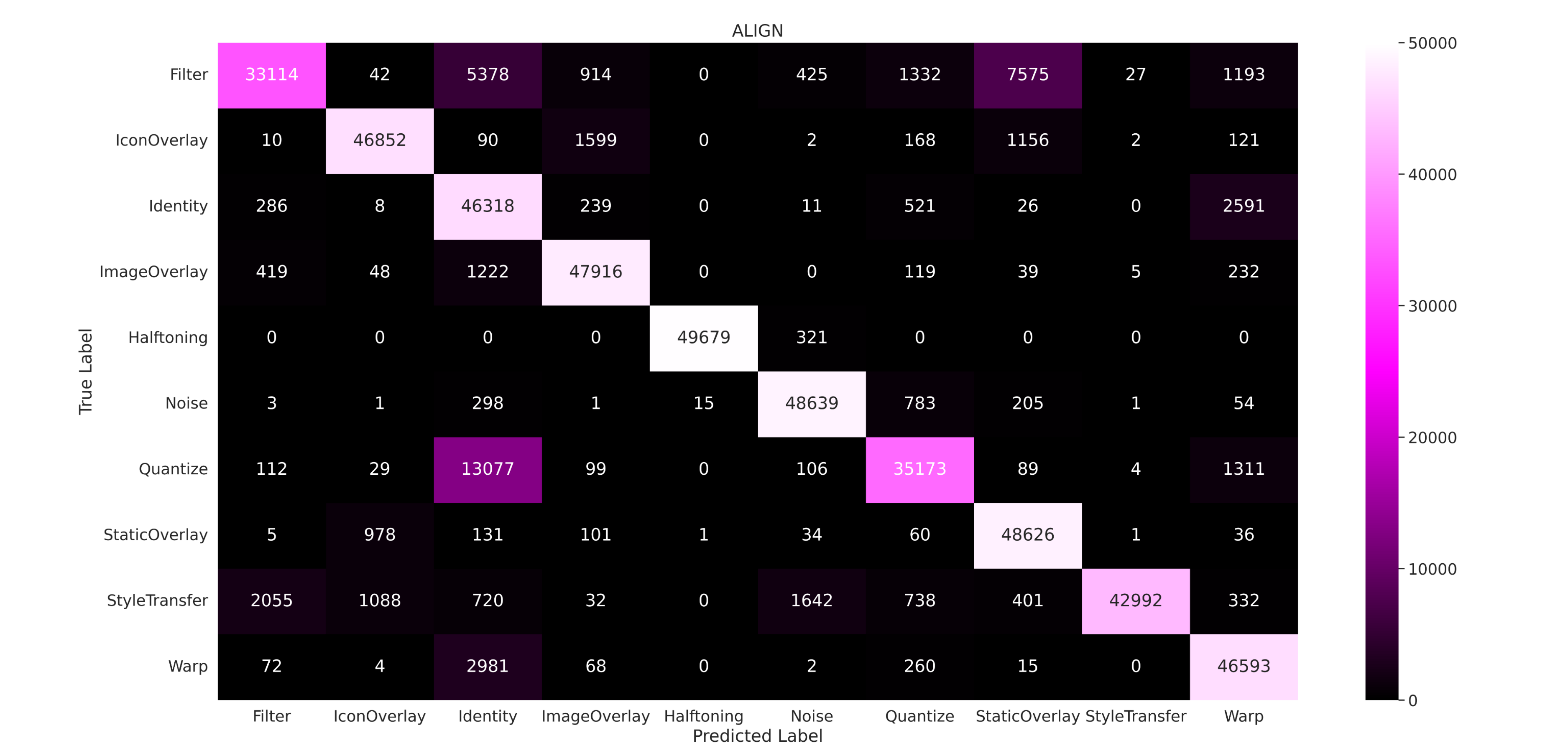}
\caption{ALIGN coarse-grained confusion matrix. Rows correspond to the 50k images of each of the 10 labels, and columns correspond to the number of predictions from the model for each of the 10 categories.}
\label{fig:align-confusion-coarse}
\vskip -0.2in
\end{figure*}

\begin{figure*}[ht]
\centering
\includegraphics[width=0.8\textwidth]{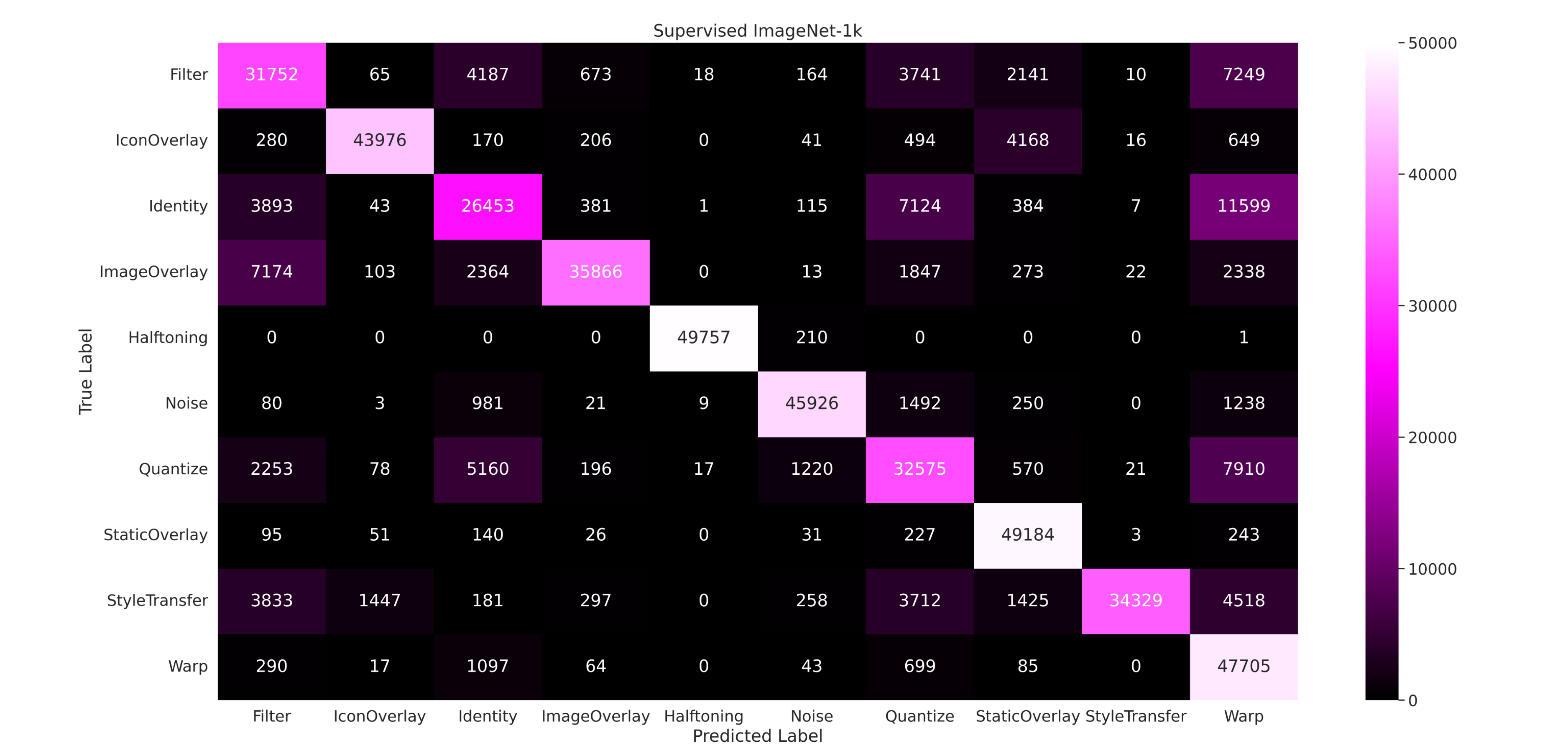}
\caption{Supervised (ImageNet-1k) coarse-grained confusion matrix. Rows correspond to the 50k images of each of the 10 labels, and columns correspond to the number of predictions from the model for each of the 10 categories.}
\label{fig:supervised-confusion-coarse}
\vskip -0.2in
\end{figure*}

\end{document}